\documentclass[10pt,journal,compsoc]{IEEEtran}
\ifCLASSOPTIONcompsoc
  \usepackage[nocompress]{cite}
\else
  \usepackage{cite}
\fi
\ifCLASSINFOpdf
\else
\fi
\usepackage{amsmath}              
\usepackage{multirow,booktabs}   
\usepackage{algorithm}           
\usepackage{algpseudocode}     
\usepackage[colorlinks, linkcolor=blue, anchorcolor=blue, citecolor=blue]{hyperref}
\usepackage{graphicx}         
\usepackage{subfigure}        

\begin{document}
\title{Adversarial Motorial Prototype Framework for Open Set Recognition}

\author{Ziheng~Xia, Penghui~Wang*, Ganggang~Dong, and~Hongwei~Liu*,~\IEEEmembership{Member,IEEE}
\IEEEcompsocitemizethanks{
\IEEEcompsocthanksitem The authors are with National Lab of Radar Signal Processing, Xidian University, No. 2 South Taibai Road, Xi’an, Shaanxi 710071, P.R. China.\protect \\
E-mail: \{xiaziheng@stu.,~wangpenghui@mail.,~dongganggang@,~hwliu@\}xidian.edu.cn.
}
\thanks{(Corresponding authors: Hongwei Liu, Penghui Wang.)}
}

\markboth{Journal of \LaTeX\ Class Files,~Vol.~14, No.~8, August~2015}%
{Shell \MakeLowercase{\textit{et al.}}: Bare Demo of IEEEtran.cls for Computer Society Journals}

\IEEEtitleabstractindextext{
\begin{abstract} % less than 200.
Open set recognition is designed to identify known classes and to reject unknown classes simultaneously. Specifically, identifying known classes and rejecting unknown classes correspond to reducing the empirical risk and the open space risk, respectively. First, the motorial prototype framework (\textbf{MPF}) is proposed, which classifies known classes according to the prototype classification idea. Moreover, a motorial margin constraint term is added into the loss function of the MPF, which can further improve the clustering compactness of known classes in the feature space to reduce both risks. Second, this paper proposes the adversarial motorial prototype framework (\textbf{AMPF}) based on the MPF. On the one hand, this model can generate adversarial samples and add these samples into the training phase; on the other hand, it can further improve the differential mapping ability of the model to known and unknown classes with the adversarial motion of the margin constraint radius. Finally, this paper proposes an upgraded version of the AMPF, \textbf{AMPF++}, which adds much more generated unknown samples into the training phase. In this paper, a large number of experiments prove that the performance of the proposed models is superior to that of other current works.
\end{abstract}

}

\maketitle
\IEEEdisplaynontitleabstractindextext
\IEEEpeerreviewmaketitle
\IEEEraisesectionheading{\section{Introduction}}

\IEEEPARstart{W}{ITH} the development of artificial intelligence technology in recent years, the application of deep learning has been pervasive in many aspects of life, such as image recognition and speech recognition\cite{face-recognition,speech-recognition}. Generally, most of the recognition studies have focused on closed set recognition (\textbf{CSR}), whose test set and training set have the same classes of data. However, in practical applications, due to the complexity of actual use scenarios, the classes of the test set might not be completely consistent with the classes of the training set. This kind of target recognition, which could contain a large number of unknown classes in the test set, is called open set recognition (\textbf{OSR})\cite{1-vs-set}.

The key to OSR is to identify unknown classes while recognizing known classes. In other words, in addition to CSR, OSR also needs to make the distribution of the unknown classes' embedding features not coincide with that of known classes. CSR corresponds to the reduction of the empirical risk, and avoiding the overlap of unknown and known classes' embedding features corresponds to the reduction of the open space risk\cite{1-vs-set}. Therefore, OSR must reduce the empirical risk and the open space risk simultaneously.

\begin{figure*}[ht]
\centering
\subfigure[Softmax(test set)] {\label{Softmax test data visualization}     
\includegraphics[width=0.38\columnwidth]{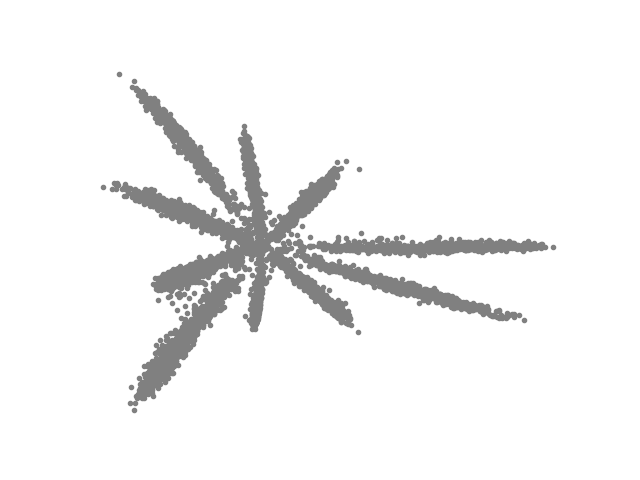}  }
\subfigure[GCPL(test set)] {\label{GCPL test data visualization}     
\includegraphics[width=0.38\columnwidth]{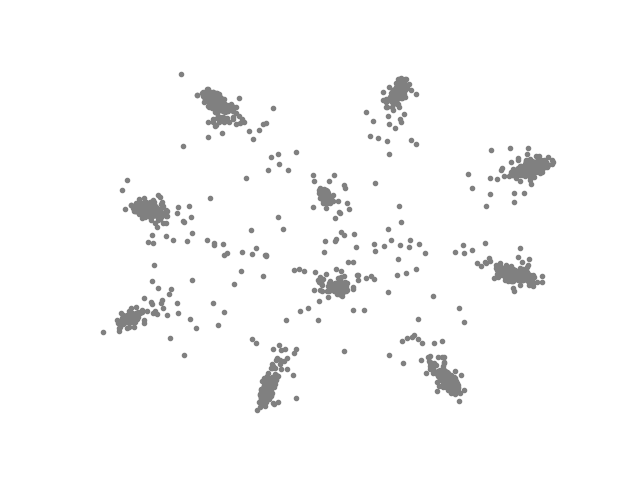}  }
\subfigure[RPL(test set)] {\label{RPL test data visualization}     
\includegraphics[width=0.38\columnwidth]{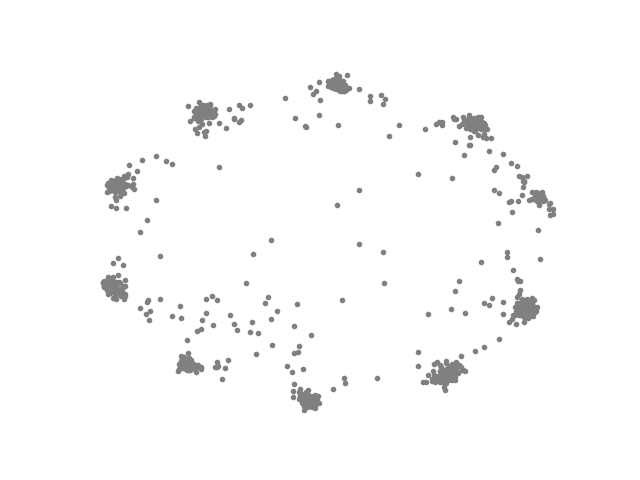}  }
\subfigure[ARPL(test set)] { \label{ARPL test data visualization}     
\includegraphics[width=0.38\columnwidth]{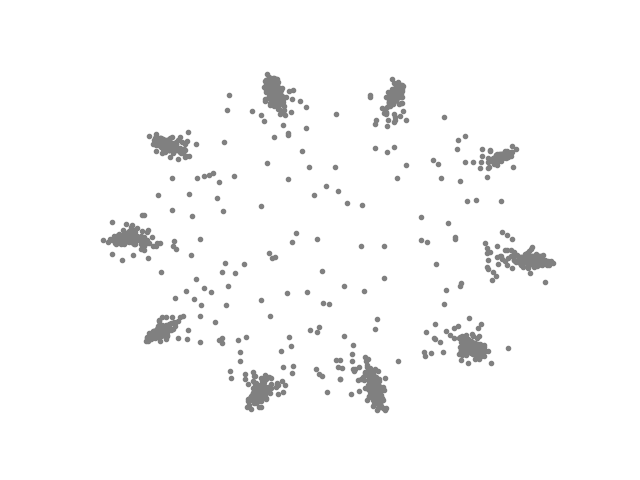}   } 
\subfigure[MPF(test set)] { \label{MPF test data visualization}     
\includegraphics[width=0.38\columnwidth]{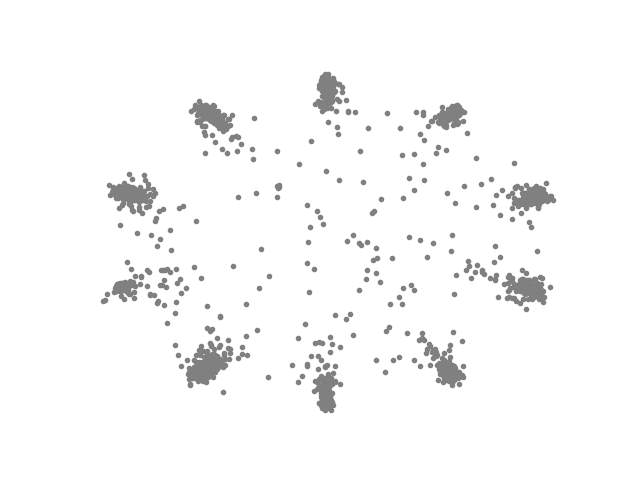}   }

\subfigure[Softmax(open set)] {\label{Softmax open set visualization}     
\includegraphics[width=0.38\columnwidth]{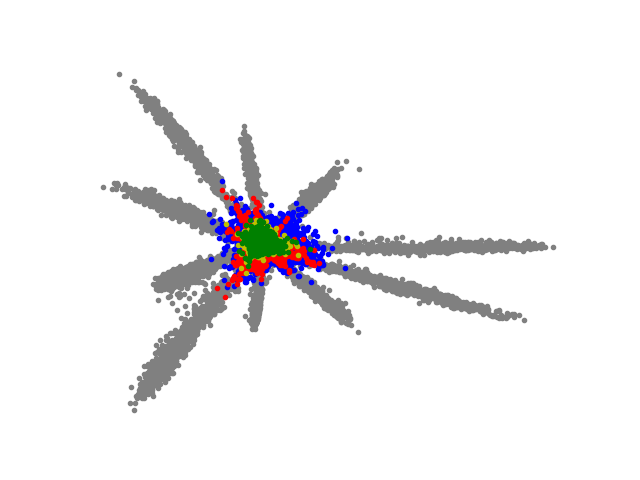}  }
\subfigure[GCPL(open set)] {\label{GCPL open set visualization}     
\includegraphics[width=0.38\columnwidth]{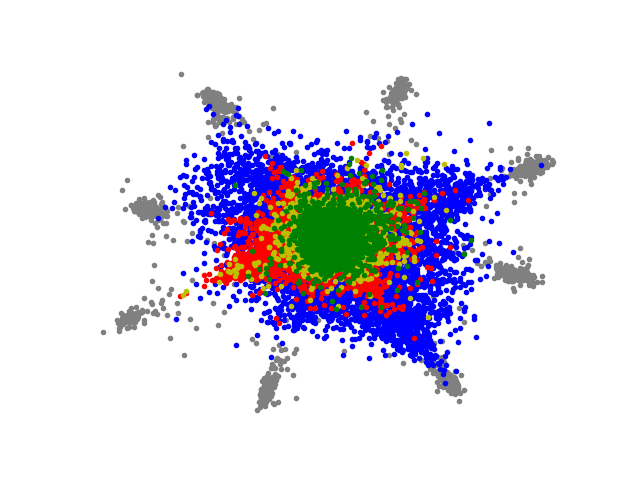}  }
\subfigure[RPL(open set)] {\label{RPL open set visualization}     
\includegraphics[width=0.38\columnwidth]{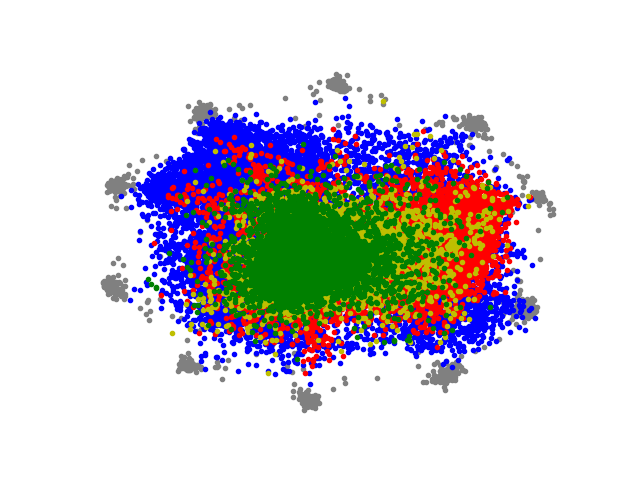}  }
\subfigure[ARPL(open set)] {\label{ARPL open set visualization}     
\includegraphics[width=0.38\columnwidth]{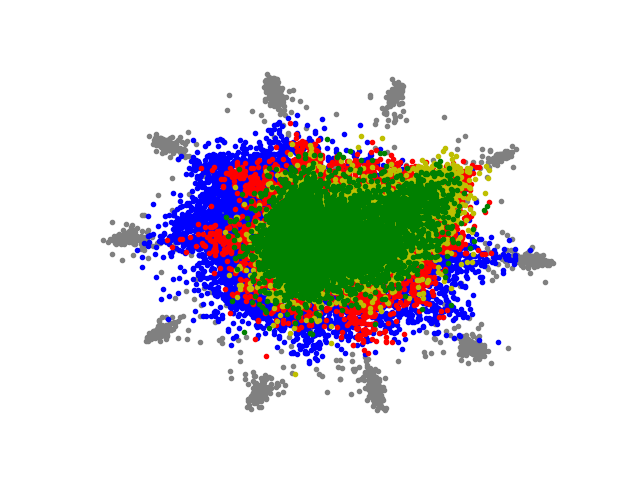}  } 
\subfigure[MPF(open set)] {\label{MPF open set visualization}     
\includegraphics[width=0.38\columnwidth]{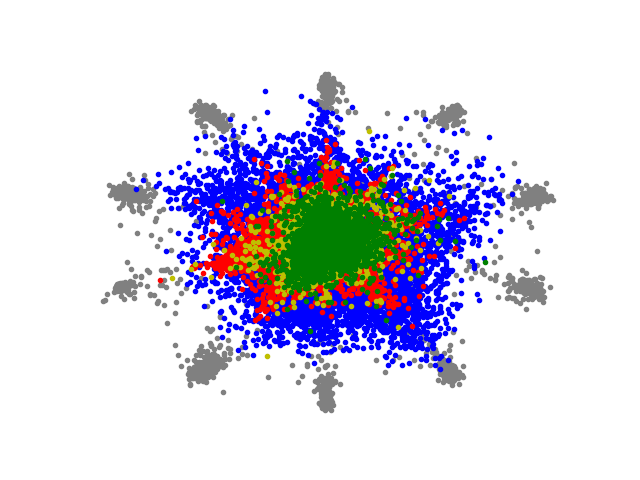}  }

\caption{\textbf{The visualization results of LENET++ on known and unknown classes}\cite{objectosphere,ARPL,RPL,GCPL-conference}. In this figure, MNIST (gray) is used for training the LENET++ network, while KMNIST (blue), SVHN (red), CIFAR10 (green) and CIFAR100 (yellow) are used for open set evaluation. The $5$ columns in the figure correspond to the visualization results of the LENET++ network under $5$ different methods, which are softmax, GCPL, RPL, ARPL and our MPF. The first row in the figure shows the visualization results on the MNIST test set under different methods. According to the second row of the figure, our MPF can effectively reduce the open space risk, while the features of the unknown classes have some extent of overlap with those of known classes under other methods.}
\label{the 1st fig}
\end{figure*}

When a classical deep neural network addresses empirical risk, a fully connected layer is connected to the end of the network for classification, and softmax is used to train the network. As shown in Fig. \ref{Softmax test data visualization}, the feature space will be divided into several half-open spaces by hyperplanes, and the features of the known classes will be forced to fall into the corresponding half-open space. Although the network trained by softmax can classify MNIST effectively, Fig. \ref{Softmax open set visualization} proves that this network cannot avoid the overlap of embedding features between known and unknown classes.

To avoid the open space risk caused by the use of the softmax training network, some prototype methods are proposed for OSR\cite{GCPL-conference,RPL,ARPL,GCPL-journal}. Among them, H. M. Yang et al. proposed the GCPL model\cite{GCPL-conference}. This model utilizes a prototype to represent each known class in the feature space and forces the features of the training data to be close to the corresponding prototypes. Compared with softmax, the advantage of GCPL is that it can not only reduce the empirical risk effectively but also improve the compactness of each class cluster in the feature space, which is helpful to reduce the open space risk. However, because GCPL does not care about the position of the prototypes in the feature space, it can be seen from Fig. \ref{GCPL open set visualization} that there are still two known clusters that are overlapping with unknown classes.

Guangyao Chen et al. proposed the RPL/ARPL model for OSR\cite{RPL, ARPL}. Different from GCPL, these two models encourage the features of the training data to be far away from the corresponding reciprocal points. According to Fig. \ref{RPL test data visualization} and \ref{ARPL test data visualization}, both two models can reduce the empirical risk effectively. In terms of the open space risk, it can be seen from Fig. \ref{RPL open set visualization} that there is still a small number of unknown classes that overlap with known classes in the feature space, while ARPL is better than RPL. Although the performance of ARPL is already very good, as shown in Fig. \ref{ARPL open set visualization}, the distribution of the three unknown classes represented by red, yellow and green is not tight, and it tends to be distributed outwards. As unknown classes become more complex and diverse, it will create open space risk. We believe that the potential risks of this model may be related to its counterintuitive classification ideas. 

To better reduce the two risks, this paper proposes a novel motorial prototype framework (\textbf{MPF}). On the basis of the prototype classification idea, MPF adds a motorial margin constraint term into the loss function, and this term can further compress the distribution range of the known classes in the feature space. As shown in Fig. \ref{MPF test data visualization} and \ref{MPF open set visualization}, compared with other models, MPF can reduce the empirical risk and the open space risk effectively at the same time.

When facing the endless unknown classes in the test phase, it is obviously not sufficient to reduce the open space risk according to only the limited known classes. Therefore, many studies generate unknown class data and add them into the training phase to further reduce the open space risk\cite{ARPL,G-OpenMax,OSRCI,osr-review-88,osr-review-90}. Considering this idea, an "adversarial motion" properties model, the adversarial motorial prototype framework (\textbf{AMPF}), is proposed on the basis of MPF. Combining the known class data with the generated data in the training phase, this model forces the unknown class features to be far away from known classes with the adversarial motion of the margin constraint radius. In addition, an enhanced version of the AMPF model, \textbf{AMPF++}, is proposed, and this model can add the generated unknown class data of more regions in the open space into the training phase to further improve the OSR performance.

Our contributions mainly focus on the following:
\begin{enumerate}
\item A novel MPF model is proposed, and this model reduces the empirical risk and open space risk simultaneously by compressing the distribution range of known classes in the feature space;
\item On the basis of MPF, AMPF is proposed. In this model, a novel optimization strategy with adversarial motion properties is introduced, and it can effectively reduce the open space risk;
\item An enhanced version of AMPF, AMPF++, is proposed, and it can further improve the OSR performance by seeing more unknown class data generated in the training phase;
\item Many experiments conducted on the larger and more difficult ImageNet datasets, which demonstrates that our methods outperform previous approaches and achieves state-of-the-art performance.
\end{enumerate}

\section{Related Work}
\subsection{Open Set Recognition}
OSR was first defined by Walter Scheirer et al.\cite{1-vs-set}. Chuanxing Geng et al. summarized the work of OSR in detail, which mainly divides the work of OSR into two categories, including discriminative models and generative models\cite{osr-review}.

Discriminative model: The detection of unknown classes in OSR can be regarded as a binary classification. This type of binary classification function is mainly realized based on traditional machine learning methods at the beginning, especially support vector machines, such as \cite{1-vs-set, w-svm, p_i-svm, SVM-64, EPCC}. With the development of deep learning technology, many methods based on deep neural networks have been developed. Dhamija et al. evaluated the open space risk of trained deep neural networks at the earliest, and they proposed a novel objectosphere loss function to reduce the open space risk by maximizing the entropy of unknown classes\cite{objectosphere}. Bendale et al. proposed the OpenMax model, replacing the softmax layer in the deep network with the OpenMax layer for OSR\cite{OpenMax}. Rozsa et al. evaluated the classification robustness of the network trained by OpenMax and softmax and found that the robustness of the OpenMax model was easily affected by some adversarial images\cite{osr-review-76}. Hassen et al. trained the network according to the criterion that the embedding features of the same class are closer to each other and proved that this method has a statistically significant performance improvement\cite{osr-review-77}. There are two important studies based on the reconstruction idea in which the reconstruction error of the known classes is smaller than that of the unknown classes, namely, classification-reconstruction for OSR (CROSR) and class conditioned auto-encoder for OSR (C2AE), and both of them have very good performance on a variety of data \cite{CROSR,C2AE}. Different from the above work, ShuLei et al. used the distance function of a submodel to detect the potential category information in the unknown classes\cite{osr-review-84}.

In these works, various complex strategies are carried out to make the embedding features more discriminative. However, these models, which are only constructed based on known classes, are obviously limited in their ability to identify unknown classes.

Generative model: To strengthen the OSR performance, a generative model adds the generated unknown class data into the training process. Therefore, the method of generating data is especially important. At present, the mainstream data generation method is the generative adversarial network (GAN) proposed by Goodfellow Ian et al\cite{GAN}. On the basis of the OpenMax model, ZongYuan Ge et al. combined the characteristics of GAN and proposed the G-OpenMax model\cite{G-OpenMax}. As a performance benchmark of the generative model, it can effectively detect unknown classes in handwritten digital sets, while it fails to perform well on natural images. Different from G-OpenMax, Neal Lawrence et al. combined the GAN structure with the idea of an encoder-decoder and proposed the OSR with counterfactual images (OSRCI) model\cite{OSRCI}, and it has influenced many subsequent studies, such as \cite{osr-review-88,osr-review-90}. Different from the generation mechanism of GAN, Yang Yu et al. proposed the adversarial sample generation model, which can generate not only unknown classes but also known classes to augment the training data \cite{ASG}. Sun X et al. proposed conditional Gaussian distribution learning (CGDL) on the basis of a variational autoencoder, and this model forces different latent features to approximate different Gaussian models for OSR\cite{CGDL}. Zhang Hongjie et al. proposed Hhbrid including an encoder, a classifier and a flow-based density estimator, in which the density estimator is used to detect whether a sample belongs to an unknown class\cite{Hhbrid}. 

These models have improved OSR performance because they have seen a lot of unknown class data in the training phase. Compared with discriminative model, the main idea of generative model plays a crucial role in improving the OSR performance. Therefore, it can be expected that more such methods will emerge in the future.

\subsection{Prototype Learning}
The prototype is usually used to refer to one or more points that can represent the cluster\cite{NPC}. The best known classification method by prototype is k-NN. 
On the basis of k-NN, Kohonen et al. proposed the learning vector quantization (LVQ) model, which allocates one or more prototypes for each class of data to represent and distinguish different classes in data\cite{LVQ}. On the basis of LVQ, many studies have been devoted to enhancing the performance of this model. Some designs more upgrade rules in the training stage, while others design novel loss functions for parameter optimization\cite{LVQ,GCPL-33,GCPL-36,GCPL-37,GCPL-38}. Furthermore, Liu, C. L. et al. proposed three algorithms based on parameter optimization and obtained the best recognition results on multiple handwritten character datasets\cite{GCPL-42}. Most of these early research results were based on the manual design of features. It was not until the maturity of deep learning technology that the end-to-end and powerful nonlinear mapping ability of deep neural networks integrated feature extraction and prototype learning. As a result, many prototype learning methods based on deep neural networks have been developed, such as in \cite{prototypical-network,improved-prototypical-network,prototypical-network-3}. GCPL in Fig. \ref{GCPL test data visualization} is also a prototype learning model designed based on a deep network, which compares the classification results under the optimization of various loss functions\cite{GCPL-conference}. Subsequently, this model was modified to a convolutional prototype network (CPN) and achieved good performance on a variety of data for OSR\cite{GCPL-journal}.

For the OSR, the methods based on prototype focus on reducing the intra-class distance of the known classes, and it usually ignores the potential risks posed by unknown classes, causing the open space risk.

\section{Motorial Prototype Framework}
\subsection{Problem Definition}
Given a set of training data $D_{tr} = \{ (x_1, y_1), (x_2, y_2),\dots \}$ with $N$ known classes, $y_i \in L_{tr}=\{1,\dots,N\}$ is the label of data $x_i$. The potential unknown data are denoted by $D_{u}$, whose label is considered to be $N+1$. It is likely that the potential unknown data could come from considerably different classes, and their specific classes are not important to OSR; thus, they are place in the same category $N+1$ here. In the test phase, there is a large quantity of test data, $D_{te}=\{s_1,s_2,\dots\}$, whose label belongs to $\{1,\dots,N,N+1\}$. In the OSR problem, there is no doubt that test data $D_{te}$ will simultaneously include the potential unknown data $D_{u}$ and the test data of known classes $D_{te}-D_u$. Moreover, all of the data come from the $d$-dimensional full space $\mathbf{R}^d$.

The goal of OSR is to minimize the empirical risk and the open space risk simultaneously, which can be formulated as
\begin{equation}
\arg \min_f  \big\{R_e(D_{tr}, L_{tr};f)+\epsilon \cdot R_o(D_{te};f) \big\},
\end{equation}
where $R_e$ is the empirical risk, $R_o$ is the open space risk, $\epsilon$ is a positive regularization parameter, and $f : \mathbf{R}^d \mapsto \mathbf{N}$ is a multiclass recognition function. It is obvious that the function $f$ should map the original data $x$ to one of the labels $\{1, 2, \dots, N\}$ when $x$ belongs to the known classes and map the data $x$ to the label $N+1$ when $x$ belongs to the unknown classes. Hence, the open space risk $R_o(D_{te};f)$ can be further formulated as
\begin{equation} \label{open space risk}
R_o(D_{te};f)=\frac{\int_{D_{te}-D_u} f(x) \mathrm{d} x}{\int_{D_u} f(x) \mathrm{d} x}.
\end{equation}

As a result, all that is needed in OSR is to determine an embedding function $f$ based on the training data $D_{tr}$ and training label $L_{tr}$, and this function not only could classify the known classes $D_{te}-D_u$ with high accuracy but also could detect the unknown classes $D_u$ as far as possible.

\subsection{Prototype for Classification}
Some studies have set multiple prototypes for each cluster\cite{LVQ,GCPL-journal}, which makes it possible to make the cluster distribution in the feature space not tight enough, thus increasing the open space risk. Therefore, this paper sets only one prototype center for a certain cluster, namely, the prototype center $O=\{O^k, k=1,2,\cdots,N\}$. For any original data $x$, the probability that its label $y$ belongs to class $k$ can be calculated by the following formula:
\begin{equation}
p(y=k|x,\Theta,O)=\frac{e^{-d(\Theta(x), O^k)}}{\Sigma_{i=1}^N e^{-d(\Theta(x), O^i)}},
\end{equation}
where $\Theta$ is the embedding function of the original data $x$, and $d(\Theta(x), O^k)$ is the distance between $\Theta(x)$ and $O^k$.To ensure that the distribution of $N$ prototype centers is not too scattered, $N$ prototype centers are randomly initialized by Gaussian distributions. The embedding features should be closer to the corresponding prototype center, which is used as a criterion to train the network. Its loss function can be denoted as
\begin{equation}
L_c(x;\theta, O)=-log\ p(y=k|x,\Theta,O), \label{loss function Lc}
\end{equation}
where $\theta$ is the network parameter.

\begin{figure}[ht] 
\centering
\includegraphics[width=0.95\columnwidth]{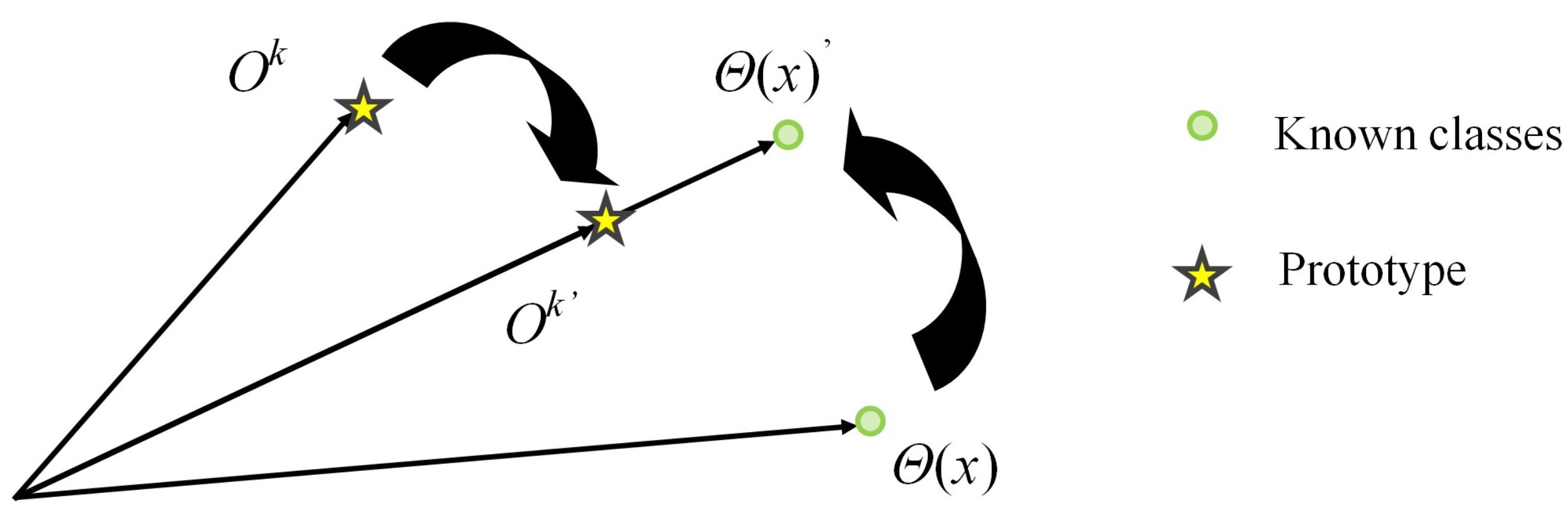}   
\caption{\textbf{Distance optimization diagram}. The optimization of the network will reduce the distance $d(\Theta(x), O^k)$, and its reduction means decrease in $d_e(\Theta(x), O^k)$ and increase in $d_d(\Theta(x), O^k)$, which will lead to the reduction of spatial distance between $\Theta(x)$ and $O^k$, and the elimination of the vector angle between $\Theta(x)$ and $O^k$, respectively.}
\label{distance setting}
\end{figure}

For distance setting, like ARPL, it is also considered insufficient to characterize the Euclidean distance between the embedding feature and the prototype center in the optimization process. Therefore, the distance setting of ARPL is also used in this paper. Specifically, $d(\Theta(x), O^k)$ can be expressed as
\begin{equation}
\begin{split}
d (\Theta(x),O^k)  &=d_e(\Theta(x),O^k)-d_d(\Theta(x),O^k), \\
d_e(\Theta(x),O^k) &=\frac{1}{m}||\Theta(x)-O^k||_2^2, \\
d_d(\Theta(x),O^k) &=\Theta(x) \cdot O^k,                  
\end{split}
\end{equation}
where $m$ is the dimension of the $\Theta(x)$ and $O^k$. When the network is optimized, $\Theta(x)$ and $O^k$ will be adjusted as shown in Fig \ref{distance setting}. The optimization of $L_c(x;\theta, O)$ is close to convergence only if the vectors $\Theta(x)$ and $O^k$ are approximately a straight line and $\Theta(x)$ and $O^k$ is very close. This distance setting ensures that the category clusters are as outwardly distributed as possible, which can reduce the open space.

\subsection{Motorial Margin Constraint}
Training the network only according to Eq. \eqref{loss function Lc} has a limited effect on reducing the open space risk. This paper hopes to reduce the open space risk by further compressing the distribution range of the known classes in the feature space. Therefore, a motorial margin constraint term $L_o$ is proposed, and it can be specifically expressed as
\begin{equation}
L_o(x;\theta, O, R) = \max\{0, d_e(\Theta(x^k), O^k)-R\}, \label{loss function Lo}
\end{equation}
where $x^k$ represents the training data with label $k$.

\begin{figure*}[!t] 
\centering
\includegraphics[width=1.8\columnwidth]{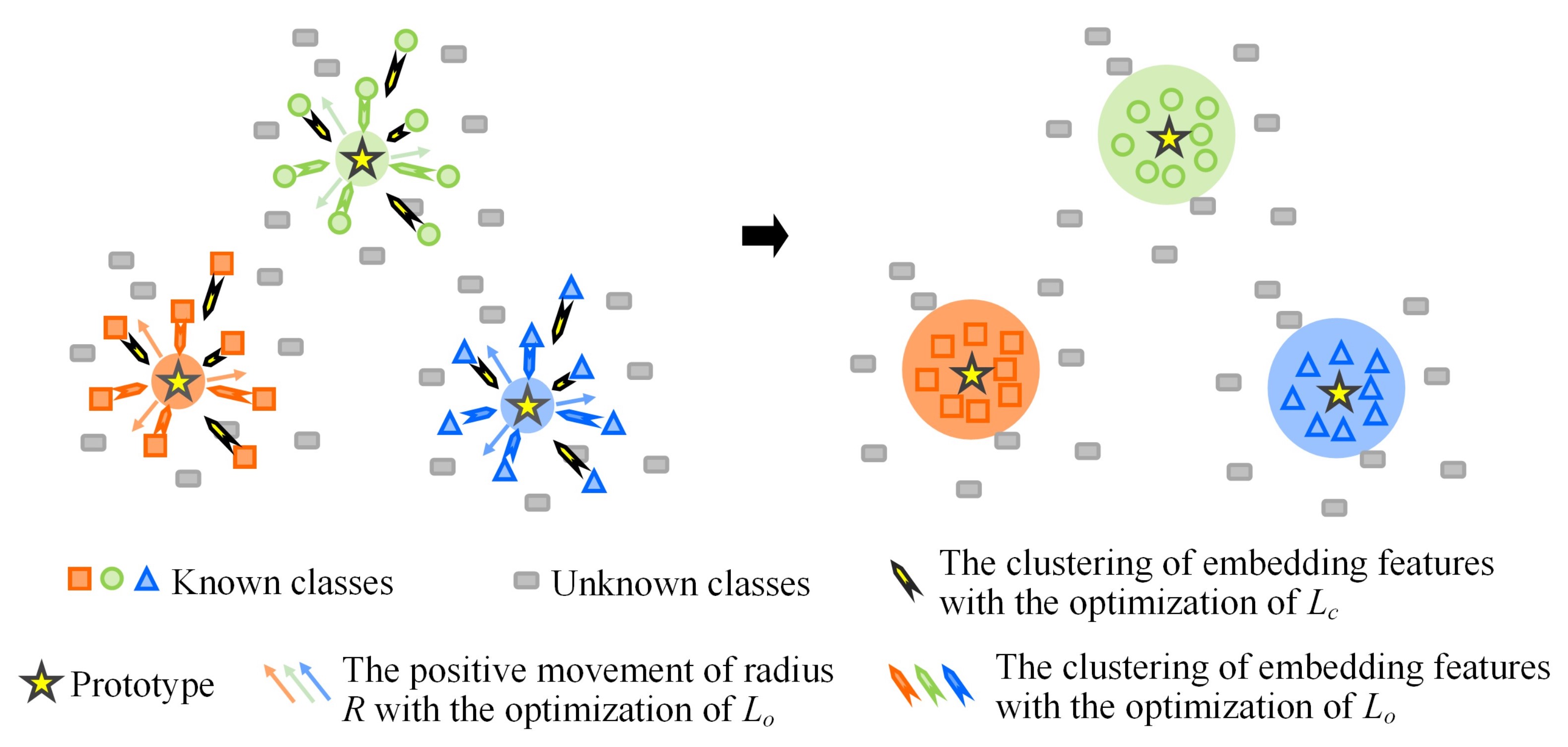}   
\caption{\textbf{MPF model optimization diagram}. On the one hand, the MPF model makes the embedding features of the known data surround the corresponding prototype closely with the optimization of the loss function $L_c$ and $L_o$; on the other hand, the optimization of the loss function $L_o$ is supposed to increase the radius $R$ until $L_o=0$.}
\label{MPF diagram}
\end{figure*}

This constraint initializes a learnable parameter $R$ with a value of $0$. Since $d_e(\Theta(x^k), O^k)$ is always nonnegative, the value of $Lo$ is equal to $d_e(\Theta(x^k), O^k)-R$ in the initial stage of the network optimization process. When optimizing the network in the direction of the negative gradient of the loss function, $R$ will gradually increase, and $d_e(\Theta(x^k), O^k)$ will gradually decrease. The reduction in $d_e(\Theta(x^k), O^k)$ means that the aggregation degree of each embedding feature towards its respective prototype center is further enhanced. Therefore, the loss function $L_o$ can assist $L_c$ in further strengthening the embedding capability of the network. This enhancement can not only further reduce the empirical risk but also further reduce the open space risk by reducing the distribution range of the known classes in the feature space. As shown in Fig. \ref{MPF diagram}, the optimization of this constraint term will eventually make the embedding features of each known class fall into a hypersphere with respective prototype center $O^k$ and radius $R$.

When the radius $R$ increases with the optimization of the network, it is said to be in "positive motion". As the radius $R$ increases and $d_e(\Theta(x^k), O^k)$ decreases, there must be a time that $R$ will be greater than $d_e(\Theta(x^k), O^k)$, and then, $L_o$ will always be equal to $0$. At this point, $L_o$ fails, and the radius $R$ no longer increases, which is said to be the end of the "positive motion".

In the optimization of $L_o$, the margin constraint radius $R$ has a "positive motion" in which $R$ only increases but does not decrease. In Section $4$, the radius $R$ will not only increase under the optimization of $L_o$ but also decrease under the optimization of other loss functions. The radius $R$ will show an adversarial reciprocating movement of increasing and decreasing, and the ability of the network to identify unknown classes will be enhanced greatly.

\subsection{Motorial Prototype Framework for Open Set Recognition}
Combining the loss function $L_c$ and $L_o$, the optimization of the whole network is expressed by the following loss function:
\begin{equation}
L(x,y;\theta, O, R) = L_c(x;\theta, O)+\lambda L_o(x;\theta, O, R), \label{loss fuction L}
\end{equation}
where $\lambda\in(0,1)$ is the proportionality coefficient that controls the weight of the motorial margin constraint term. The network model based on Eq. \eqref{loss fuction L} is called the motorial prototype framework(\textbf{MPF}) in this paper. When the training process of this model is completed, a network with trained parameter $\theta$, prototype center $O$ and radius $R$ can be obtained.

In the initial stage of the network optimization, the motion of radius R can be expressed as
\begin{equation}
R^{t+1}=R^t-\mu^t \cdot \frac{\partial L^t}{\partial R^t} = R^t+\mu^t\lambda,
\end{equation}
where $\mu$ is the learning rate of the network and $t$ is the iteraiton number. Therefore, the radius $R$ is in "positive motion" with the rate $\mu^t\lambda$.

According to the visualization results in Fig. \ref{MPF test data visualization}, MPF can effectively reduce the empirical risk. At the same time, it can be seen from Figs. \ref{Softmax open set visualization}, \ref{GCPL open set visualization}, \ref{RPL open set visualization}, \ref{ARPL open set visualization} and \ref{MPF open set visualization} that MPF can reduce the open space risk more effectively than softmax, GCPL and RPL, even better than ARPL.

\section{Adversarial Enhancement of Motorial Prototype Framework}
When the distribution of unknown classes becomes increasingly complex in real tests, MPF will most likely no longer perform, as shown in Fig. \ref{MPF open set visualization}. Specifically, the model does not use any prior information of unknown classes, which limits the performance of the model. Therefore, many methods choose to use GAN to generate some unknown samples\cite{GAN} and add these generated samples into the training phase to further reduce the open space risk, such as \cite{ARPL,G-OpenMax,OSRCI}.

In this paper, a novel network optimization strategy based on MPF, the adversarial motorial prototype framework (\textbf{AMPF}), is proposed by referring to these generation models. On the one hand, this strategy sets up a generator different from the traditional GAN, which can generate a large number of adversarial samples and adds these samples into the training process. On the other hand, this strategy injects a new "adversarial motion" mode into the motion of the radius $R$, and the OSR performance will be further improved with this motion of the radius $R$.

\subsection{Generator and Discriminator}
GAN consists of a generator $G$ and a discriminator $D$. The generator $G$ can map a prior distribution $Z$ to data $G(Z)$ in $\mathbf{R}^d$ space. The discriminator $D$ maps the input data to $0$ or $1$, and it can be used to identify whether the input is real or generated data.

Given the distribution $\{z_1,z_2,\cdots,z_n\}$ and the training data $\{x_1,x_2,\cdots,x_n\}$, discriminator $D$ is optimized by
\begin{equation} \label{optimize D in AMPF}
\max_D \frac{1}{n}\sum_{i=1}^n\Big[\log D\big(x_i\big)+\log\big(1-D(G(z_i))\big)\Big],
\end{equation}
in such a way that it can better distinguish real data from generated data. To generate data that can fool discriminator $D$, generator G is optimized by
\begin{equation}
\max_G \frac{1}{n}\sum_{i=1}^n\log D\big(G(z_i)\big).
\end{equation}

In fact, the training process of the GAN is essentially a game process between the generator $G$ and discriminator $D$. The final result of the game is that the generator $G$ can generate data $G(z_i)$, which is very similar to the distribution of real data $x_i$.

Similar to the composition of GAN, AMPF includes a generator $G$ and a discriminator $D$. In addition, AMPF contains a classifier $C$. Classifier $C$ is the network represented by parameter $\theta$ in the MPF model, and its embedding function is denoted as $\Theta$ here.

\begin{figure}[ht] 
\centering
\includegraphics[width=1.0\columnwidth]{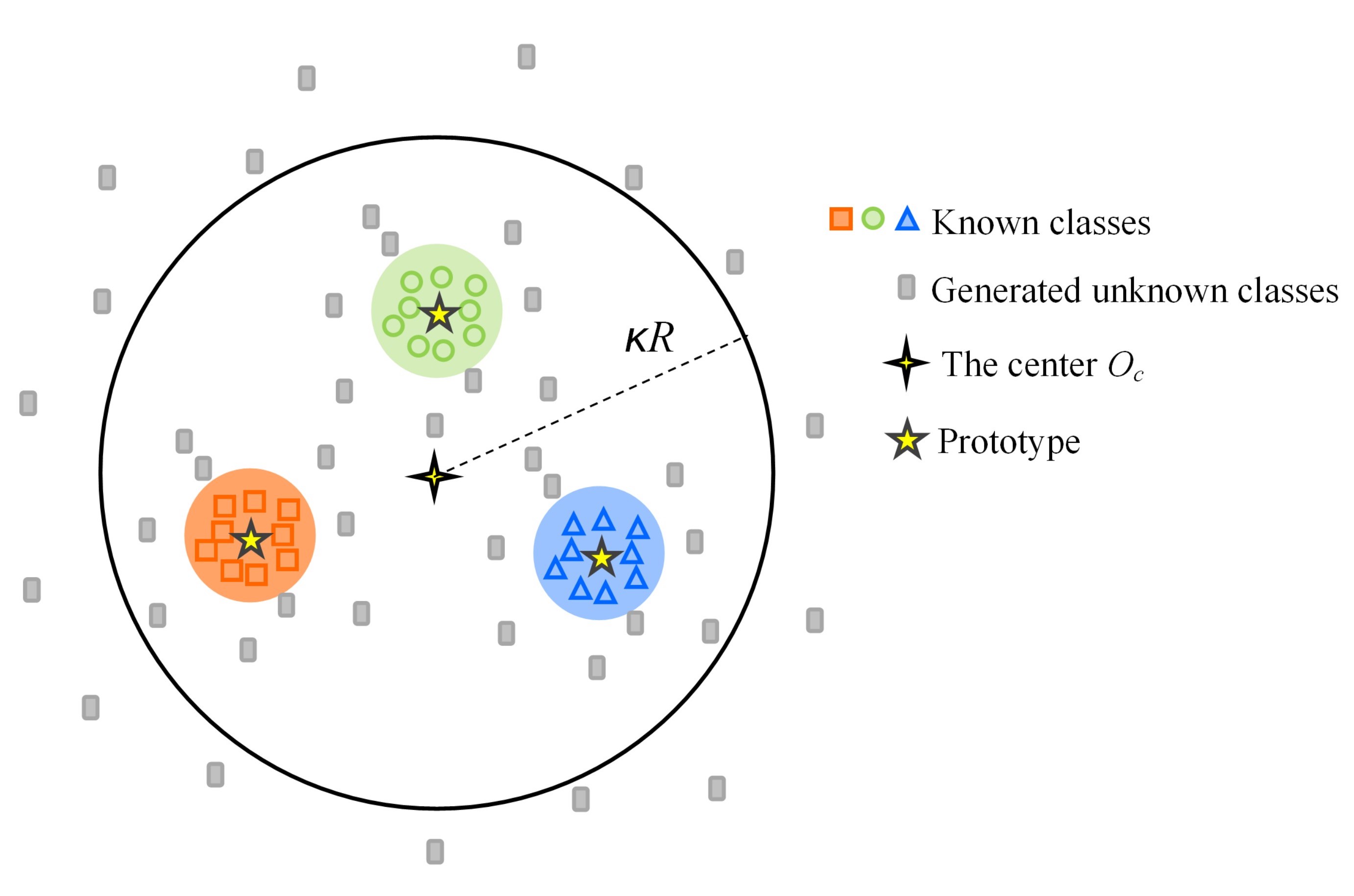}   
\caption{\textbf{The generated data distribution diagram in the AMPF model}. The AMPF model can generate not only data similar to the known data but also data at the edge of the open space region.}
\label{AMPF_1 diagram}
\end{figure}

\begin{figure*}[ht]
\centering
\includegraphics[width=1.8\columnwidth]{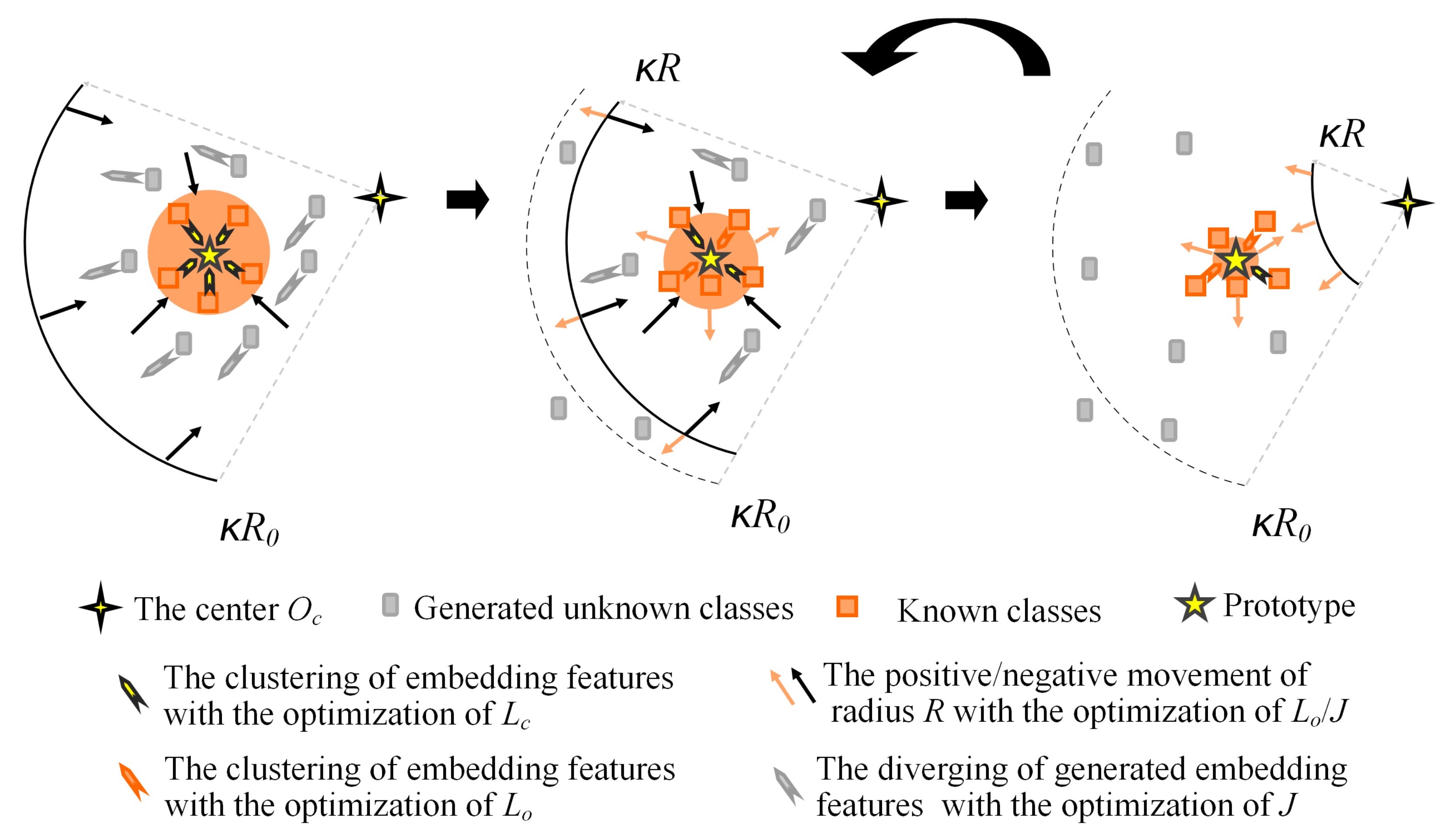}   
\caption{\textbf{Diagram of adversarial motion in the AMPF model}. Under the optimization of the loss functions $L_o$ and $J$, the radius R moves in the positive and negative directions, respectively. As the radius R increases, the optimization of the loss function $L_o$ will help $L_c$ to make the known category features cluster much more closely. As the radius R decreases, the optimization of the loss function $J$ forces the generated data features to reach the edge of the open space.}
\label{AMPF_2 diagram}
\end{figure*}

As shown in Fig. \ref{AMPF_1 diagram}, this paper hopes that the generator $G$ in AMPF can also generate data in the open space far away from the known classes prototypes, which can be achieved through the following optimization:
\begin{equation} \label{generate far away data}
\min_G \frac{1}{n}\sum_{i=1}^n\max\Big\{ 0, \kappa R - d_e\big(\Theta(G(z_i)), O_c\big) \Big\},
\end{equation}
where $\kappa$ is the hyperparameter, and $O_c$ is the mean value of all prototype centers $O^k$,
which can be calculated by $O_c=\frac{1}{N}\sum_{k=1}^NO^k$.

The summation part of Eq. \eqref{generate far away data} is denoted as
\begin{equation}
J(z_i;O_c)=\max\Big\{ 0, \kappa R - d_e\big(\Theta(G(z_i)), O_c\big) \Big\}.
\end{equation}
Then, the optimization strategy of generator $G$ can be summarized as
\begin{equation} \label{optimize G in AMPF}
\max_G \frac{1}{n}\sum_{i=1}^n\Big[\log D\big(G(z_i)\big)-\alpha J(z_i;O_c)\Big],
\end{equation}
where $\alpha\in(0,1)$ is the hyperparameter, and it is used to control the weight of the generating data far from all of the prototype centers.

Finally, the joint optimization of Eq. \eqref{optimize D in AMPF} and \eqref{optimize G in AMPF} will enable generator $G$ to generate data distributed around the known classes prototype and outside the $\kappa R$ distance centered on $O_c$. Obviously, it can be seen that $\kappa\gg1$. For convenience, the space beyond the distance $\kappa R$ with $O_c$ as the center of the circle is called the edge region of the open space.

\subsection{Adversarial Motorial Margin Constraint Radius}
In this paper, it is hoped that classifier $C$ can complete the clustering of the known classes and map all of the generated data to the edge region of the open space at the same time. By strengthening the differential mapping ability of classifier $C$, the OSR performance of the model can be enhanced effectively.

The optimization strategy for training the differential mapping ability of classifier $C$ is as follows:
\begin{equation} \label{optimize C in AMPF}
\min_C \frac{1}{n}\sum_{i=1}^n\Big[L(x_i,y_i;\theta, O, R)+\beta J(z_i;O_c) \Big],
\end{equation}
where the hyperparameter $\beta\in(0,1)$ controls the weight at which classifier $C$ maps the generated data to the edge region of the open space.

When classifier $C$ is optimized according to the negative gradient direction, the radius $R$ is also "moving", and its motion formula is as follows:
\begin{equation} \label{R motion in AMPF}
R^{t+1}=R^t-\mu^t \cdot \frac{\partial (L^t+\beta J^t)}{\partial R^t} = R^t+\mu^t(\lambda-\beta\kappa).
\end{equation}

If the loss function $L_o$ has failed, the motion formula of $R$ should be:
\begin{equation} \label{R pure negative motion in AMPF}
R^{t+1}=R^t-\mu^t\beta\kappa.
\end{equation}

According to $\kappa\gg1$ and $\lambda,\beta\in(0,1)$, it is easy to pick the right value in such a way that the value of $\lambda-\beta\kappa$ is less than $0$. After the radius $R$ increases to $R_0$ in a "positive motion" under the optimization of Eq. \eqref{loss fuction L}, the motion represented by Eq. \eqref{R pure negative motion in AMPF} will occur under the optimization of Eq. \eqref{optimize C in AMPF}. At this point, the radius $R$ starts to decrease, and "negative motion" occurs. To ensure that this "negative motion" can be effectively started, parameter $\kappa$ should meet the following conditions:
\begin{equation} \begin{split} \label{value of k in equation}
&\kappa R_0 > d_0+\gamma R_0 \quad \\
\Longrightarrow &\kappa >\frac{d_0}{R_0} + \gamma,
\end{split} \end{equation}
where hyperparameter $\gamma\ge1$, and $d_0$ is the sum of the distances between all prototype centers $O^i$ and the center $O_c$, which can be calculated by
\begin{equation}
d_0=\sum_{i=1}^N d_e(O^i, O_c).
\end{equation}

As shown in Fig. \ref{AMPF_2 diagram}, the optimization of the loss function $L_c$ continuously clusters the embedding features of the known classes towards the respective prototype center, which is independent of the motion of the radius $R$. In addition, the radius $R$ will enter "negative motion" starting from $R_0$ under the optimization of the loss function $J$. In this process, the optimization of the loss function $J$ will make the classifier $C$ map the generated data to the edge region of the open space. At the same time, due to the reduction in radius $R$, the loss function $L_o$ is activated, and it will assist the loss function $L_c$ to enhance the ability of classifier $C$ to cluster known classes. When the radius $R$ decreases to a certain extent, the loss function $J$ fails, and the "negative motion" ends. Then, the radius $R$ begins to enter "positive motion" under the optimization of the loss function $L_o$ until the loss function $J$ restarts the "negative motion". In other words, the radius $R$ goes into the adversarial reciprocating motion.

\begin{figure}[ht]
\centering
\includegraphics[width=0.9\columnwidth]{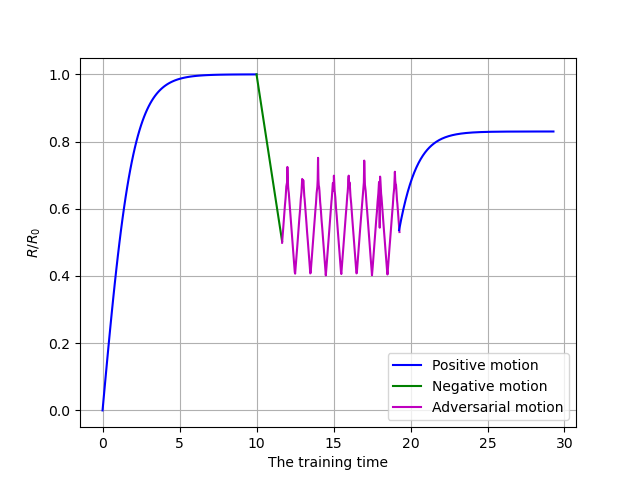}   
\caption{\textbf{The prediction diagram of radius $R$ motion trail}. This figure should be understood in conjunction with Figs. \ref{MPF diagram} and \ref{AMPF_2 diagram}. The blue curve in this figure corresponds to the optimization process shown in Fig. \ref{MPF diagram}, and the green and pink curve correspond to the optimization process shown in Fig. \ref{AMPF_2 diagram}.}
\label{R motion trail pre}
\end{figure}

\subsection{AMPF Working Procedure}
Finally, the training process of the AMPF model combining the generator $G$, discriminator $D$ and classifier $C$ is shown in Alg. \ref{AMPF algorithm}. In each training epoch, classifier $C$ is first updated according to the loss function $L$ of the MPF model, aiming to make the radius $R$ enter "positive motion" and reach the position $R_0$. Then, within each training batch, discriminator $D$, generator $G$, and classifier $C$ are updated according to the optimization strategy of the AMPF model.

The reasons for assigning the parameter $\kappa$ to the formula in step \ref{value of k in AMPF} of Alg. \ref{AMPF algorithm} are as follows:
\begin{enumerate}
\item It can satisfy Eq. \eqref{value of k in equation};
\item As the number of iterations increases, parameter $\kappa$ can enhance the "negative motion" rate, which is conducive to strengthening the differential mapping ability of classifier $C$;
\item With the increase in the number of iterations, the model gradually tends to converge, as does the value of $R_0$. Therefore, it is expected that $\kappa$ will also converge gradually.
\end{enumerate}

\begin{algorithm}
\caption{The AMPF algorithm}\label{AMPF algorithm}
\begin{algorithmic}[1]
\Require Training data $\{x_i\}$. Initialized parameters $\theta_G$ of the generator, $\theta_D$ of the discriminator and $\theta_C$ of the classifier with prototype center parameters $O$ and radius parameter $R$ in the loss layers. Hyper-parameter $\lambda,\alpha,\beta,\gamma$ and variable parameter $\kappa$. The total number of iteration $\mathrm{max\_epoch}$.

\For{$t \gets 0, \mathrm{max\_epoch}-1$}

    \State Update the classifier parameters $\theta_C$ with $O$ and $R$ by minimizing MPF loss, such as \label{optimize classifier}
    \begin{equation*}
    \nabla_{\theta_C} L(x,y;\theta, O, R).
    \end{equation*}

    \State Record the initial value of the antagonistic motion of the radius: \label{step 3 in AMPF algorithm}
      \begin{equation*}
      R_0 \gets R.
      \end{equation*}

    \For{$l \gets 0, \mathrm{max\_batch}-1$}
      \State Calculate the total distance between each category of prototype and center $O_c$: \label{step 5 in AMPF algorithm} 
      \begin{equation*}
      d_0 \gets \sum_{j=1}^N d_e(O^j,O_c).
      \end{equation*}

      \State Calculate the expansion factor $\kappa$ of the radius $R$: \label{value of k in AMPF}
      \begin{equation*}
      \kappa \gets (\gamma+\frac{d_0}{R_0}) \log(t+3).
      \end{equation*}
      
      \State Sample $\{z_1,z_2,\cdots,z_n\}$ from prior $P_{pri}(z)$ and known samples $\{(x_1,y_1),(x_2,y_2),\cdots,(x_n,y_n)\}$. \label{step 7 in AMPF algorithm}

      \State Update the discriminator parameters $\theta_D$ by:
      \begin{equation*}
      \nabla_{\theta_D} -\frac{1}{n}\sum_{i=1}^n\Big[\log D\big(x_i\big)+\log\big(1-D(G(z_i))\big) \Big].
      \end{equation*}

      \State Update the generator parameters $\theta_G$ by:
      \begin{equation*}
      \nabla_{\theta_G} -\frac{1}{n}\sum_{i=1}^n\Big[\log D\big(G(z_i)\big)-\alpha J(z_i;O_c)\Big].
      \end{equation*}

      \State Update the classifier parameters $\theta_C$ with $O$ and $R$ by: \label{optimize R adversarial motion}
      \begin{equation*}
      \nabla_{\theta_C} \frac{1}{n}\sum_{i=1}^n\Big[L(x_i,y_i;\theta, O, R)+\beta J(z_i;O_c)\Big].
      \end{equation*}
    \EndFor \label{step 11 in AMPF algorithm}

    \If{$t = \mathrm{max\_epoch}-1$}
      \State Update the classifier again by repeating step \ref{optimize classifier}.
    \EndIf
\EndFor
\end{algorithmic}
\end{algorithm}

Finally, Fig. \ref{R motion trail pre} predicts the trail of the radius $R$ in the AMPF model. The "positive motion" trajectory represented by the first blue curve corresponds to the training process shown in Fig. \ref{MPF diagram}, and the $R$ increases to the position of $R_0$ at this time. Under the optimization of the strategy that corresponds to step \ref{optimize R adversarial motion}, the "negative motion" stops when the $R$ decreases from the position of $R_0$ to a certain extent, and this stage corresponds to the green curve. Subsequently, the $R$ enters "positive motion" only under the optimization of the loss function $L_o$; and so on and so forth, the $R$ goes into an adversarial reciprocating pattern, which corresponds to the pink curve. As shown in the second blue curve in Fig. \ref{R motion trail pre}, the $R$ will enter the next motion cycle with the beginning of the next training iteration. With the adversarial movement of the $R$, the differential mapping ability of classifier $C$ to known class data and generated data will be greatly enhanced.

\section{Improvement of the Adversarial Motorial Prototype Framework}
Theoretically, the classifier of AMPF can effectively identify known and unknown classes by looking for their distribution boundaries in the feature space. However, because the test data and training data have difficulty strictly meeting an independent identical distribution, the generalization ability of the model to the test set is limited. As a result, the feature distribution of the test data is always looser than that of the training data, which can be verified in the first two rows of Fig. \ref{the 1st fig}. In other words, this limited generalization ability will create open space risk to some extent. Specifically, the unknown class embedding features within a certain distance from the central $O_c$ can overlap with known class embedding features. In fact, this phenomenon is also observed in Fig. \ref{MPF open set visualization}.

Therefore, to improve AMPF, the \textbf{AMPF++} model is proposed. It adds another generator to generate the junction data between the center $O_c$ and all the prototype centers based on the existing AMPF model structure. These generated data are also added into the adversarial training process to further enhance the differential mapping ability of the classifier to the known and unknown classes.

\subsection{Another Generator}
Given the distribution $\{z_1,z_2,\cdots,z_n\}$, the data generated by the newly added generator $G_2$ are denoted as $G_2(z_i)$. The optimization strategy of generator $G_2$ is expressed by the following formula:
\begin{equation} \label{optimize G2 in AMPF++}
\min_{G_2} \mathrm {MSE} \big\{ O_c+\delta x, \Theta\big(G_2(z_i)\big) \big\},
\end{equation}
where $\delta x$ is an error vector in the feature space, and it can be used to control the distribution of the generated data in the feature space.

This paper chooses to use $\Theta(G_2(z_i))$ approximating $O_c+\delta x$ to generate data instead of directly using $O_c+\delta x$ as the generated data. Because the former can replace the latter, the latter cannot replace the former. At the initial stage of optimization, the distribution of $\Theta(G_2(z_i))$ can be significantly different from that of $O_c+\delta x$, which will not affect the performance but will improve the differential mapping ability of the classifier because it generates more region data in the open space. At the later stage of optimization, Eq. \eqref{optimize G2 in AMPF++} gradually converges, and $\Theta(G_2(z_i))$ can approximate $O_c+\delta x$ effectively, which will achieve the desired purpose of the generation.

Next, we need to discuss how $\delta x$ can be used to control the distribution of features that we want to generate. Let the dimension of the feature space be $m$; then, $\delta x$ can be further expressed as $\delta x=(\delta x_1,\delta x_2,\cdots,\delta x_m)$.

For the convenience of derivation and calculation, this paper assumes that each component of $\delta x$ is independently distributed in a Gaussian distribution with a mean of $0$ and a variance of $\sigma^2$:
\begin{equation} \label{delta x_i}
\delta x_i \stackrel{i.i.d} \sim N(0, \sigma^2).
\end{equation}

Under the premise that $\Theta(G_2(z_i))$ can approximate $O_c+\delta x$ well, it is necessary to carefully consider the distance distribution from $O_c+\delta x$ to $O_c$ to determine the parameter $\sigma$ in $\delta x$:
\begin{equation} \begin{split}
d_e(O_c+\delta x, O_c) &= \frac{1}{m} ||\delta x||^2_2 \\
&=\frac{1}{m} \sum_{i=1}^m \delta x_i^2. \\
\end{split} \end{equation}

According to probability theory, the following expressions can be obtained: $\mathrm{E}\delta x_i^2=\sigma^2$ and $\mathrm{E}\delta x_i^4=3\sigma^4$. Since each component of $\delta x$ is independently and identically distributed in the Gaussian distribution, the square of each component of $\delta x$ is also independent and identically distributed, specifically,
\begin{equation}
\delta x_i^2 \stackrel{i.i.d} \sim 
\begin{cases}
\mathrm{E}\delta x_i^2=\sigma^2 \\
\mathrm{D}\delta x_i^2=2\sigma^4.
\end{cases}
\end{equation}

According to the central limit theorem, the sum of multiple independent identically distributed random variables approaches the Gaussian distribution. Therefore, when the dimension $m$ of the feature space is large enough, we have
\begin{equation} \begin{split}
& \frac{\sum_{i=1}^m \delta x_i^2-m\mathrm{E}\delta x_i^2}{\sqrt{m\mathrm{D}\delta x_i^2}} \rightarrow N(0,1) \\
\Longrightarrow & \sum_{i=1}^m\delta x_i^2 \rightarrow N(m\sigma^2,2m\sigma^4) \\
\Longrightarrow & \frac{1}{m} \sum_{i=1}^m\delta x_i^2 \rightarrow N(\sigma^2,\frac{2}{m}\sigma^4).
\end{split} \end{equation}

As a result, the distance from the generated data features to the central $O_c$ is approximated by a Gaussian distribution with a mean of $\sigma^2$ and a variance of $\frac{2}{m}\sigma^4$. The unknown class data characteristics that affect the model performance are distributed near the boundaries of each prototype center, and thus, we can choose the appropriate value of $\sigma$ to meet our desired generation needs.

According to the properties of the Gaussian distribution, the probability of the distribution within the range of $\sigma^2\pm3\sqrt{\frac{2}{m}}\sigma^2$ is $99.7\%$. To enable the generator to generate samples at the boundary of each prototype center and without covering each prototype center, this paper chooses the $3\sigma$ principle of Gaussian distribution to generate data, which can be denoted as follows:
\begin{equation} \begin{split} \label{sigma in delta x}
&\sigma^2(1+3\sqrt{\frac{2}{m}})=\frac{1}{N}\sum_{i=1}^N d_e(O^i, O_c) \\
\Longrightarrow &\sigma^2=\frac{1}{(1+3\sqrt{\frac{2}{m}})N}\sum_{i=1}^N d_e(O^i, O_c).
\end{split} \end{equation}

The error vector is generated by Eq. \eqref{delta x_i} and \eqref{sigma in delta x}. Then, to obtain the required generated samples, the generator $G_2$ is optimized by Eq. \eqref{optimize G2 in AMPF++}.

\subsection{AMPF++ Working Procedure}
Finally, the algorithm of the AMPF++ model is shown in Alg. \ref{AMPF++ algorithm}. Within each iteration, the AMPF algorithm is run first; then, the MPF algorithm is run, causing the radius $R$ to enter "positive motion" to increase to $R_0$; finally, the adversarial motion combined with the generator $G_2$ begins.

After the network model is trained in accordance with Alg. \ref{AMPF++ algorithm}, the probability that $x$ in the test set belongs to a known class can be determined by the following formula:
\begin{equation}
p(known|x) \propto \ exp\big(-\min_k d(\Theta(x), O^k)\big).
\end{equation}

The AMPF++ model maps the known class data to the respective prototype center and maps the unknown class data to the edge region of the entire open space to the greatest extent. Therefore, the minimum distance between the embedding feature of the test data and all of the prototype centers can be used to identify whether the test data are a known or unknown class.

\begin{algorithm}
\caption{The AMPF++ algorithm}\label{AMPF++ algorithm}
\begin{algorithmic}[1]
\Require Training data $\{x_i\}$. Initialized parameters $\theta_G, \theta_{G_2}$ of the generator and the 2nd generator, $\theta_D$ of the discriminator and $\theta_C$ of the classifier with prototype center parameters $O$ and radius parameter $R$ in the loss layers. Hyper-parameter $\lambda,\alpha,\beta,\gamma$ and variable parameter $\kappa$. The total number of iteration $\mathrm{max\_epoch}$.
\For{$t \gets 0, \mathrm{max\_epoch}-1$}
  \Procedure{AMPF}{$2\sim11$}
    \State Repeat the step: \ref{optimize classifier} to the step: \ref{step 11 in AMPF algorithm} in the AMPF algorithm.
  \EndProcedure

  \Procedure{AMPF}{$2\sim3$}
    \State Repeat the step: \ref{optimize classifier} to the step: \ref{step 3 in AMPF algorithm} in the AMPF algorithm.
  \EndProcedure

  \For{$l \gets 0, \mathrm{max\_batch}-1$}
      \Procedure{AMPF}{$5\sim7$}
        \State Repeat the step: \ref{step 5 in AMPF algorithm} to the step: \ref{step 7 in AMPF algorithm} in the AMPF algorithm.
      \EndProcedure

      \State Update the generator parameters $\theta_{G_2}$ by:
      \begin{equation*}
      \nabla_{\theta_{G_2}} \frac{1}{n}\sum_{i=1}^n \mathrm{MSE}\Big[O_c+\delta x, \Theta\big(G_2(z_i)\big)\Big].
      \end{equation*}
      \State Update the classifier parameters $\theta_C$ with $O$ and $R$ by:      
      \begin{equation*}
      \nabla_{\theta_C} \frac{1}{n}\sum_{i=1}^n\Big[L(x_i,y_i;\theta, O, R)+\beta J_2(z_i;O_c)\Big],
      \end{equation*}
      where $J_2(z_i;O_c)=\max\Big\{ 0, \kappa R - d_e\big(\Theta(G_2(z_i)), O_c\big) \Big\} $.
  \EndFor

    \If{$t = \mathrm{max\_epoch}-1$}
      \State Update the classifier again by repeating step \ref{optimize classifier} in the AMPF algorithm.
    \EndIf
\EndFor
\end{algorithmic}
\end{algorithm}

\section{Experiments}
\subsection{Experimental Settings}
\subsubsection{Datasets}
Like many other papers\cite{OpenMax,G-OpenMax,OSRCI,C2AE,CROSR,CGDL}
, this paper selects MNIST\cite{MNIST}, SVHN \cite{SVHN}, CIFAR10 \cite{CIFAR}, CIFAR+10, CIFAR+50 and TinyImageNet\cite{TinyImageNet} as the experimental data for the open set recognition task. MNIST, SVHN and CIFAR10 all contain data of $10$ categories, from which $6$ categories are randomly selected as known classes, and the remaining $4$ categories are unknown classes. For CIFAR+10 and CIFAR+50, $4$ classes are randomly selected from CIFAR10 as known classes, and then, $10$ and $50$ classes are randomly selected from CIFAR100 as corresponding unknown class data. For Tiny ImageNet, $20$ categories are randomly selected as known classes, and the remaining $180$ types of data are regarded as unknown classes.

\subsubsection{Network, Optimizer and Other Parameters}
In the model of MPF, AMPF and AMPF++, this paper uses the same convolutional neural network as \cite{OSRCI} as the classifier, which has $9$ convolutional layers and $1$ full connection layer. In addition to the TinyImageNet data using the Adam optimizer, the momentum stochastic gradient descent (SGD-M) optimizer is used to optimize the classifier for the other data sets\cite{Adam,SGD}. The initial learning rate of the network is set to $0.1$, dropping to one-tenth of the original rate every $30$ epochs.

In the AMPF and AMPF++ models, this paper uses the same generator and discriminator as \cite{cite_discriminator_generator}. Both of them use the Adam optimizer with a learning rate of $0.0002$ for optimization.

The hyperparameters $\lambda$, $\alpha$, $\beta$ and $\gamma$ involved in this paper are set to $0.1$, $0,1$, $0.1$ and $10$, respectively, and the dimension of the feature space $m$ is set to $128$.

\begin{table*}
\begin{center}
\caption{The closed set accuracy results test on variable methods and data sets. Every value is averaged among five randomized trials. The best results are indicated in bold. We reproduce the softmax, GCPL, RPL, ARPL and ARPL+CS, and we copy the remaining results from \cite{GCPL-journal}.}
\begin{tabular}{ccccccc} 
\toprule
Method & MNIST(\%) & SVHN(\%) & CIFAR10(\%) & CIFAR+10(\%) & CIFAR+50(\%) & TinyImageNet(\%) \\
\midrule
Softmax                   & 99.5$\pm$0.2 & 94.7$\pm$0.6 & 80.1$\pm$3.2 & 96.3$\pm$0.6 & 96.4$\pm$0.6 & 72.9$\pm$4.3 \\ 
OpenMax  \cite{OpenMax}   & 99.5$\pm$0.2 & 94.7$\pm$0.6 & 80.1$\pm$3.2 & - & - & - \\ 
G-OpenMax\cite{G-OpenMax} & 99.6$\pm$0.1 & 94.8$\pm$0.8 & 81.6$\pm$3.5 & - & - & - \\ 
OSRCI \cite{OSRCI}        & 99.6$\pm$0.1 & 95.1$\pm$0.6 & 82.1$\pm$2.9 & - & - & - \\ 
CROSR \cite{CROSR}        & 99.2$\pm$0.1 & 94.5$\pm$0.5 & 93.0$\pm$2.5 & - & - & - \\ 
CPN\cite{GCPL-journal}    & 99.7$\pm$0.1 & 96.7$\pm$0.4 & 92.9$\pm$1.2 & - & - & - \\
GCPL\cite{GCPL-conference}& \textbf{99.8$\pm$0.1} & 96.7$\pm$0.4 & 92.4$\pm$1.7 & 96.4$\pm$0.7 & 96.4$\pm$0.8 & 62.3$\pm$4.7 \\
RPL\cite{RPL}             & \textbf{99.8$\pm$0.1} & \textbf{96.9$\pm$0.4} & 94.6$\pm$1.7 & 96.5$\pm$0.6 & 96.6$\pm$0.6 & 62.8$\pm$3.7 \\
ARPL\cite{ARPL}           & 99.7$\pm$0.1 & 96.6$\pm$0.4 & 94.5$\pm$1.9 & 96.4$\pm$0.5 & 96.4$\pm$0.6 & 76.1$\pm$4.1 \\
ARPL+CS\cite{ARPL}        & 99.7$\pm$0.1 & 96.6$\pm$0.4 & 95.4$\pm$1.6 & 97.1$\pm$0.6 & 97.2$\pm$0.5 & 79.8$\pm$3.2 \\
\midrule
MPF                       & \textbf{99.8$\pm$0.1} & 96.7$\pm$0.4 & 94.5$\pm$1.9 & 96.5$\pm$0.7 & 96.5$\pm$0.8 & 75.4$\pm$4.8 \\
AMPF                      & \textbf{99.8$\pm$0.1} & \textbf{96.9$\pm$0.4} & 95.4$\pm$1.5 & 97.3$\pm$0.6 & 97.2$\pm$0.7 &79.5$\pm$4.3 \\
AMPF++                    & \textbf{99.8$\pm$0.1} & \textbf{96.9$\pm$0.4} & \textbf{96.0$\pm$1.5} & \textbf{97.5$\pm$0.7} & \textbf{97.4$\pm$0.6} & \textbf{81.1$\pm$3.4} \\
\bottomrule
\end{tabular}\label{closed set acc table}
\end{center}
\end{table*}

\subsubsection{Evaluation Metrics}
Similar to \cite{OpenMax,G-OpenMax,OSRCI,C2AE,CROSR,CGDL}, this paper also selects the area under the receiver operating characteristic curve (\textbf{AUROC}) to evaluate the performance of the model.
The AUROC evaluates the ability of the model to discriminate the unknown classes by ranking the predicted probability of the sample belonging to the known classes from high to low; its greatest advantage is that there is no need to specify the working threshold of the model.

Since AUROC only evaluates the ability of the model to identify unknown classes but does not evaluate the accuracy of the model to identify known classes, other indicators must be added to the experiments to evaluate the model completely. Some work evaluates the recognition ability of the model to the known classes through the F1-measure, the harmonic mean of the precision and recall, such as \cite{OpenMax, G-OpenMax,C2AE}. However, it is necessary to make a curve of the F1-measure change with the working threshold of the model to evaluate the identification ability for known classes, which is obviously inconvenient.

For convenience, this paper introduces an open set classification rate (\textbf{OSCR}) indicator that is independent of the working threshold to replace the F1-measure\cite{objectosphere}. Let $\tau$ be a score threshold. The correct classification rate (CCR) is the fraction of the samples where the correct class $k$ has maximum probability and has a probability greater than $\tau$: 
\begin{equation} \begin{split}
& CCR(\tau)= \\
& \frac{|\{x|x\in(D_{te}-D_u)\land \arg\max_k P(k|x)=\hat{k}\land P(\hat{k}|x)\ge\tau\}|}{|D_{te}-D_u|}.
\end{split} \end{equation}

The false positive rate(FPR) is the fraction of samples from unknown data $D_u$ that are classified as any known class $k$ with a probability greater than $\tau$:
\begin{equation}
FPR(\tau)=\frac{|\{x|x\in D_u\land \max_kP(k|x)\ge\tau\}|}{|D_u|}.
\end{equation}

Therefore, OSCR based on CCR and FPR is an indicator similar to AUROC, which evaluates the model by calculating the area under the corresponding curve. The larger the OSCR value of the model is, the stronger the recognition ability of the model.

\subsection{Results and Analysis for Closed Set Recognition}
In this paper, the known class introduced in Section 6.1.1 is used in CSR. Theoretically speaking, the ability of the MPF, AMPF and AMPF++ models proposed in this paper to identify unknown class data should be enhanced sequentially. The enhancement of the ability to identify unknown class data is usually built on the basis of sacrificing the ability to recognize known class data. However, it can be seen from Tab. \ref{closed set acc table} that the recognition ability to the known classes data of the three models proposed in this paper is not decreased, but is increasing; moreover, the model proposed in this paper, AMPF++, achieves the best test results in all CSR tests. As a result, all of these results fully prove the effectiveness of the model proposed in this paper.

\begin{table*}
\caption{The AUROC results of the open set recognition test on variable methods and data sets. Every value is averaged among five randomized trials. The best results are indicated in bold. We reproduce the GCPL and obtain its test results; moreover, other test results are copied from \cite{ARPL,GCPL-journal,PROSER,GDFR,CAC,CVAECap,MLOSR}.}
\begin{center}
\begin{tabular}{ccccccc}
\toprule
Method & MNIST(\%) & SVHN(\%) & CIFAR10(\%) & CIFAR+10(\%) & CIFAR+50(\%) & TinyImageNet(\%) \\
\midrule
Softmax                   & 97.8 & 88.6 & 67.7 & 81.6 & 80.5 & 57.7 \\ 
OpenMax\cite{OpenMax}     & 98.1 & 89.4 & 69.5 & 81.7 & 79.6 & 57.6 \\
G-OpenMax\cite{G-OpenMax} & 98.4 & 89.6 & 67.5 & 82.7 & 81.9 & 58.0 \\
OSRCI\cite{OSRCI}         & 98.8 & 91.0 & 69.9 & 83.8 & 82.7 & 58.6 \\
C2AE\cite{C2AE}           & 98.9 & 92.2 & 89.5 & 95.5 & 93.7 & 74.8 \\
CROSR\cite{CROSR}         & 99.1 & 89.9 & 88.3 & 91.2 & 90.5 & 58.9 \\
CGDL\cite{CGDL}           & 99.4 & 93.5 & 90.3 & 95.9 & 95.0 & 76.2 \\
CPN\cite{GCPL-journal}    & 99.0 & 92.6 & 82.8 & 88.1 & 87.9 & 63.9 \\
PROSER\cite{PROSER}       & -    & 94.3 & 89.1 & 96.0 & 95.3 & 69.3 \\
GDFR\cite{GDFR}           & -    & 95.5 & 83.1 & 92.8 & 92.6 & 64.7 \\
CAC\cite{CAC}             & 98.7 & 94.2 & 80.3 & 86.3 & 87.2 & 77.2 \\
CVAECap\cite{CVAECap}     & 99.2 & 95.6 & 83.5 & 88.8 & 88.9 & 71.5 \\ 
MLOSR\cite{MLOSR}         & 98.9 & 92.1 & 84.5 & 89.5 & 87.7 & 71.8 \\
GCPL\cite{GCPL-conference}& 99.3 & 95.1 & 85.8 & 91.9 & 89.5 & 70.0 \\
RPL\cite{RPL}             & 99.3 & 95.1 & 86.1 & 85.6 & 85.0 & 70.2 \\
ARPL\cite{ARPL}           & 99.6 & 96.3 & 90.1 & 96.5 & 94.3 & 76.2 \\
ARPL+CS\cite{ARPL}        & \textbf{99.7} & 96.7 & 91.0 & 97.1 & 95.1 & 78.2 \\
\midrule
MPF                       & 99.6 & 96.3 & 89.9 & 96.6 & 94.3 & 76.0 \\
AMPF                      & 99.6 & 96.7 & 91.3 & \textbf{97.3} & 95.1 & 79.3 \\
AMPF++                    & \textbf{99.7} & \textbf{96.8} & \textbf{91.6} & \textbf{97.3} & \textbf{95.4} & \textbf{79.7} \\
\bottomrule
\end{tabular} \label{AUROC table}
\end{center}
\end{table*}

\begin{table*}
\begin{center}
\caption{The open set classification rate(OSCR) curve results of the open set recognition test. Every value is averaged among five randomized trials. The best results are indicated in bold. All of results in this table are copied from \cite{ARPL} except our results.}
\begin{tabular}{ccccccc}
\toprule
Method & MNIST(\%) & SVHN(\%) & CIFAR10(\%) & CIFAR+10(\%) & CIFAR+50(\%) & TinyImageNet(\%) \\
\midrule
Softmax                     & 99.2 & 92.8 & 83.8 & 90.9 & 88.5 & 60.8 \\ 
GCPL\cite{GCPL-conference}  & 99.1 & 93.4 & 84.3 & 91.0 & 88.3 & 59.3 \\
RPL\cite{RPL}               & 99.4 & 93.6 & 85.2 & 91.8 & 89.6 & 53.2 \\
ARPL\cite{ARPL}             & 99.4 & 94.0 & 86.6 & 93.5 & 91.6 & 62.3 \\
ARPL+CS\cite{ARPL} & \textbf{99.5} & 94.3 & 87.9 & 94.7 & 92.9 & 65.9 \\
\midrule
MPF                         & 99.4 & 94.0 & 86.5 & 93.8 & 91.7 & 62.1 \\
AMPF                        & 99.4 & 94.3 & 88.1 & 94.9 & 93.0 & 67.8 \\
AMPF++ &\textbf{99.5}&\textbf{94.5}&\textbf{89.0}&\textbf{95.1}&\textbf{93.3}&\textbf{69.0} \\
\bottomrule
\end{tabular} \label{OSCR table}
\end{center}
\end{table*}

\subsection{Results and Analysis for Open Set Recognition}
The test results of OSR in this paper are shown in Tabs. \ref{AUROC table} and \ref{OSCR table}. The six types of datasets used in this test include an increasing number of classes of data, which means that the corresponding open set identification difficulty is also increasing. When the difficulty of OSR is not large, the performance difference between the different methods is not obvious; for the MNIST datasets, the performance of the different methods is generally good, and the highest AUROC value of $99.7$ was only $1.9\%$ higher than the lowest value of $97.8$. However, when the OSR difficulty gradually increased, the performance difference between the different methods was reflected; for the TinyImageNet datasets, the maximum AUROC value of $79.7$ is $38.4\%$ higher than the minimum value of $57.6$.

Tabs. \ref{AUROC table} and \ref{OSCR table} show that the OSR performance of the MPF, AMPF and AMPF++ models proposed in this paper increases sequentially, as theoretically designed. The performance of the AMPF and AMPF++ models with the generated data added into the training process is significantly better than that of the MPF model without the generated data. Similarly, some other methods based on generative models, such as CROSR, C2AE and ARPL+CS, are also better than some methods based on discriminative models, such as softmax and OpenMax, in identifying unknown class data. The AMPF++ model shows overwhelming advantages in the ability to identify unknown class data and known class data. In short, the OSR performance of the proposed method is generally better than that of other current methods.

\begin{figure*}[ht]
\centering
\subfigure[] {\label{R motion trail}     
\includegraphics[width=0.9\columnwidth]{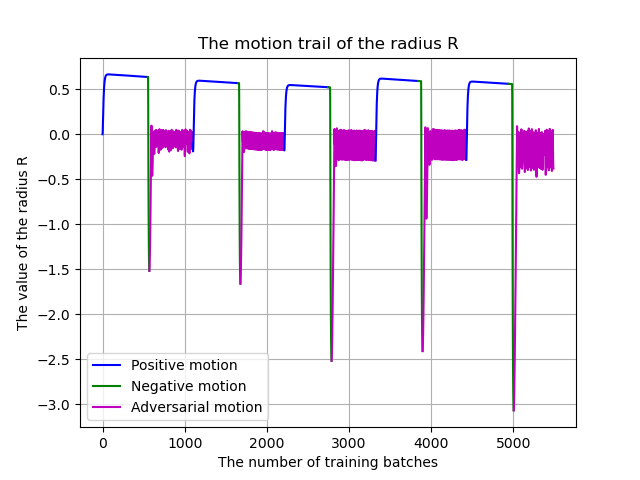}  }
\subfigure[] {\label{local R motion trail}     
\includegraphics[width=0.9\columnwidth]{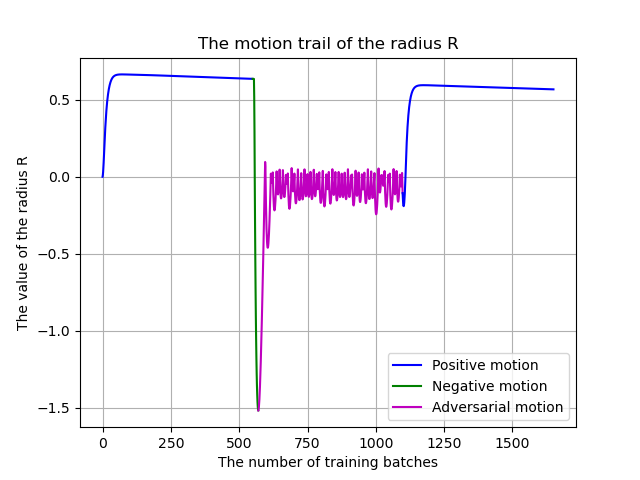}  }
\caption{\textbf{Experimental diagram of the radius $R$ motion trail}. Fig. \ref{local R motion trail} is a local view of Fig. \ref{R motion trail}. Similar to Fig. \ref{R motion trail pre}, the movement of radius R in the AMPF model experiment also presents positive, negative and adversarial motion trials.}
\end{figure*}

\subsection{The Motion of the Radius R}
This subsection gives a detailed discussion of the trajectory of radius $R$ in the real training process. Since the motion of radius $R$ in the MPF model has been included by AMPF and the motion of radius $R$ in the AMPF++ model is not fundamentally different from that of AMPF, this subsection discusses only the motion of radius $R$ in the AMPF model.

Taking MNIST as an example, the training process of the first $5$ epochs is selected to discuss the motion of radius $R$. As shown in Fig. \ref{R motion trail}, the whole figure shows $5$ periodic movements that correspond to $5$ epochs of model training. During the first period of motion, the "positive motion" corresponding to the blue curve occurs at radius $R$ starting from $0$, which corresponds to the MPF model. Then, as predicted in Fig. \ref{R motion trail pre}, there is a rapid "negative motion", followed by reciprocating "adversarial motion". Strictly speaking, "positive motion" should include only the ascending phase of the blue curve in the figure, which is not distinguished in the drawing for consistency with Fig. \ref{R motion trail pre}.

Different from the motion trajectory predicted in Fig. \ref{R motion trail pre}, the radius $R$ will decrease to a negative number in the actual training. Although the decrease in the radius $R$ to a negative number no longer has meaning in the physical image, it does not affect the activation of the loss function $L_o$; in contrast, it will prolong the working time of the loss function $L_o$, which is conducive to the clustering of the known class features. In addition, we also observed that the value of the radius $R$ decreased when the "negative motion" stopped in each period, which is related to the increase in parameter $\kappa$ with the increase in epochs. At the beginning of the next period, the radius $R$ increases to a new $R_0$, which is usually smaller than the $R_0$ of the previous period. The reason is that the clusters within the class are becoming increasingly tight, and sometimes, it will increase due to the interference of the separation point. In addition, in the second half of the blue curve, the value of the radius $R$ does not stay the same as predicted but gradually decreases at a very small rate, which should be related to the properties of the SGD-M optimizer.

\subsection{Visualization of Features in Our Models}
\begin{figure*}[ht]
\centering
\subfigure[MPF($\lambda=0$)] {\label{MPF_0_vis}  
\includegraphics[width=0.48\columnwidth]{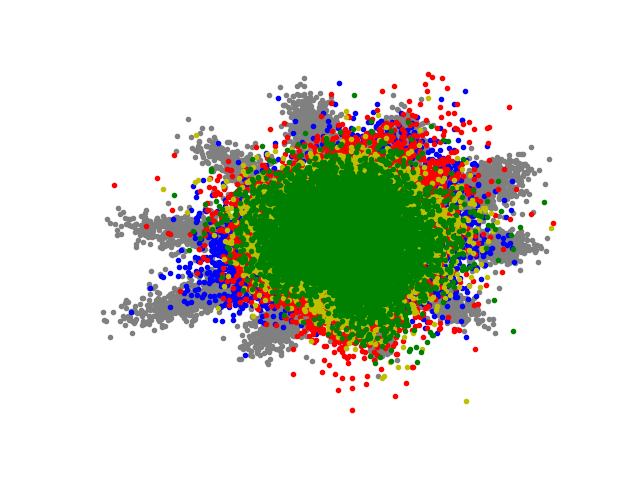}  }
\subfigure[MPF($\lambda=0.1$)] {  \label{MPF_1_vis}  
\includegraphics[width=0.48\columnwidth]{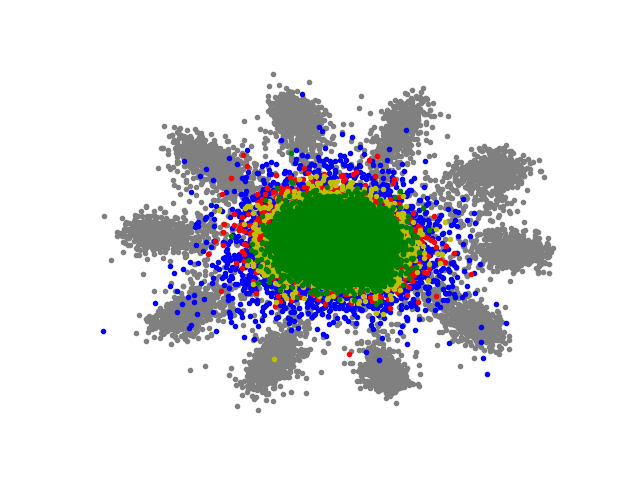}  }
\subfigure[AMPF] {   \label{AMPF_vis} 
\includegraphics[width=0.48\columnwidth]{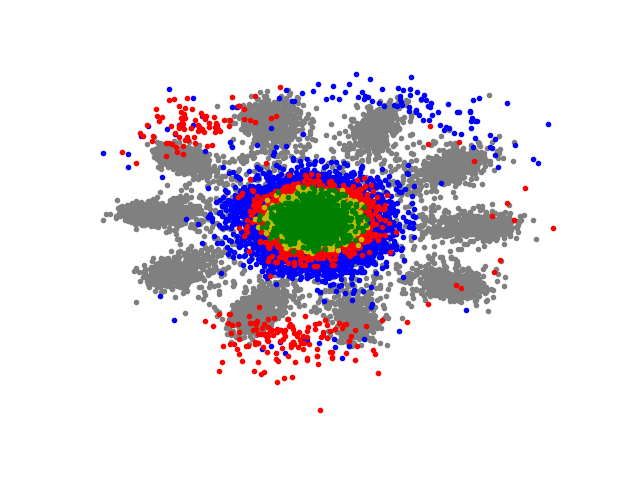}  }
\subfigure[AMPF++] {   \label{AMPF++_vis} 
\includegraphics[width=0.48\columnwidth]{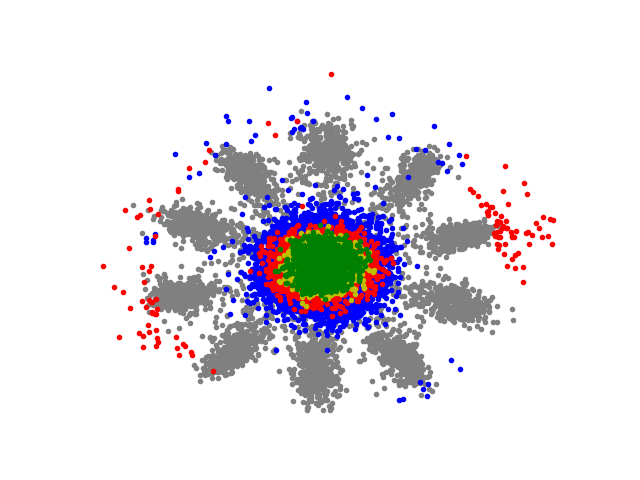}  }
\caption{\textbf{The 2D feature visualization results on known and unknown classes}. Similar to Fig. \ref{the 1st fig}, MNIST (gray) is used as a known class, and KMNIST (blue), SVHN (red), CIFAR10 (green) and CIFAR100 (yellow) are used as unknown classes for open set evaluation. To compare with Fig. \ref{MPF_1_vis}, the value of hyperparameter $\lambda$ is also set to $0.1$ in Figs. \ref{AMPF_vis} and \ref{AMPF++_vis}.}
\label{the 7th fig}
\end{figure*}

To better compare the three models proposed in this paper, two types of visualization methods are used on these models in this section. In the subsection "2D Features Visualization", the output of the network feature vector is set to two; thus, we can directly plot the features on the 2D surface for visualization. Subsection "t-SNE Visualization" uses the same convolutional neural network as \cite{OSRCI}, and the feature dimension of this network is $128$. In this subsection, the t-SNE method is used to visualize the known and unknown embedding features.

\begin{figure*}[ht]
\centering
\subfigure[MPF] {  \label{MPF_inner}  
\includegraphics[width=0.65\columnwidth]{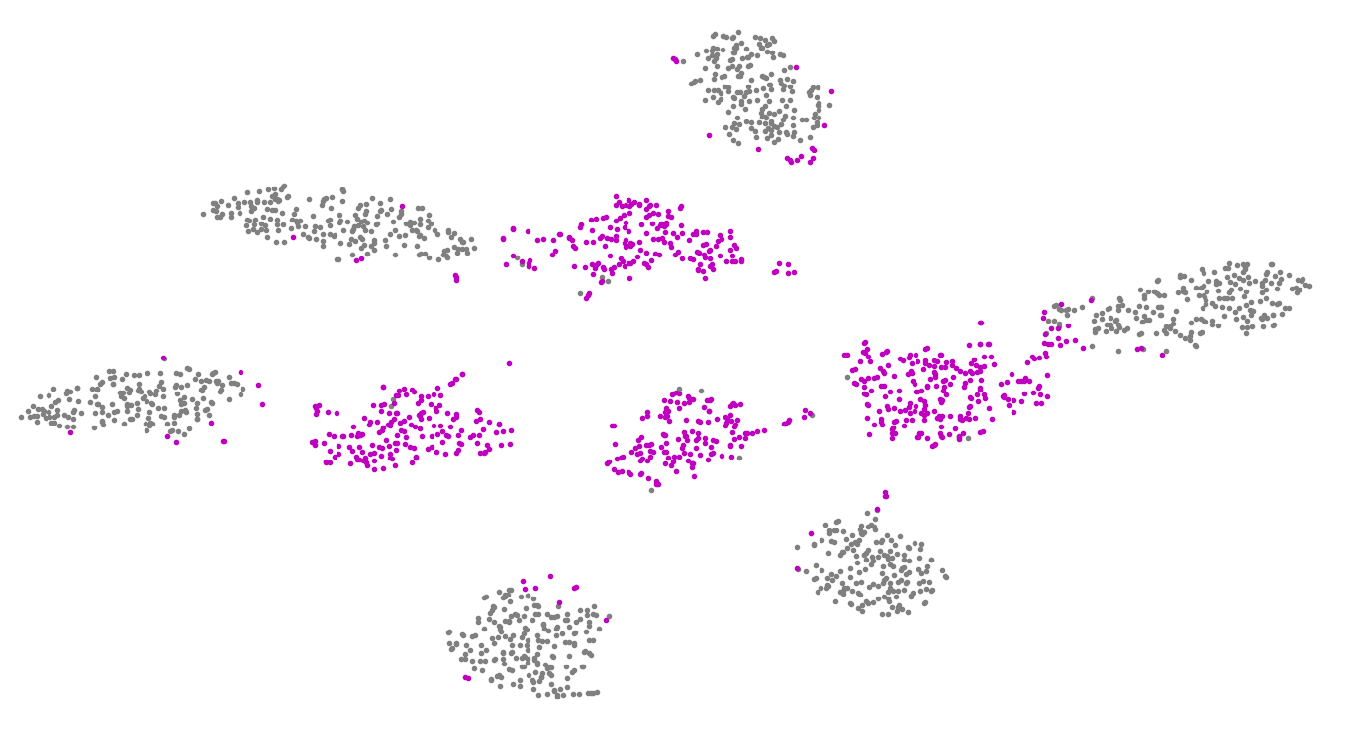}  }
\subfigure[AMPF] {   \label{AMPF_inner} 
\includegraphics[width=0.65\columnwidth]{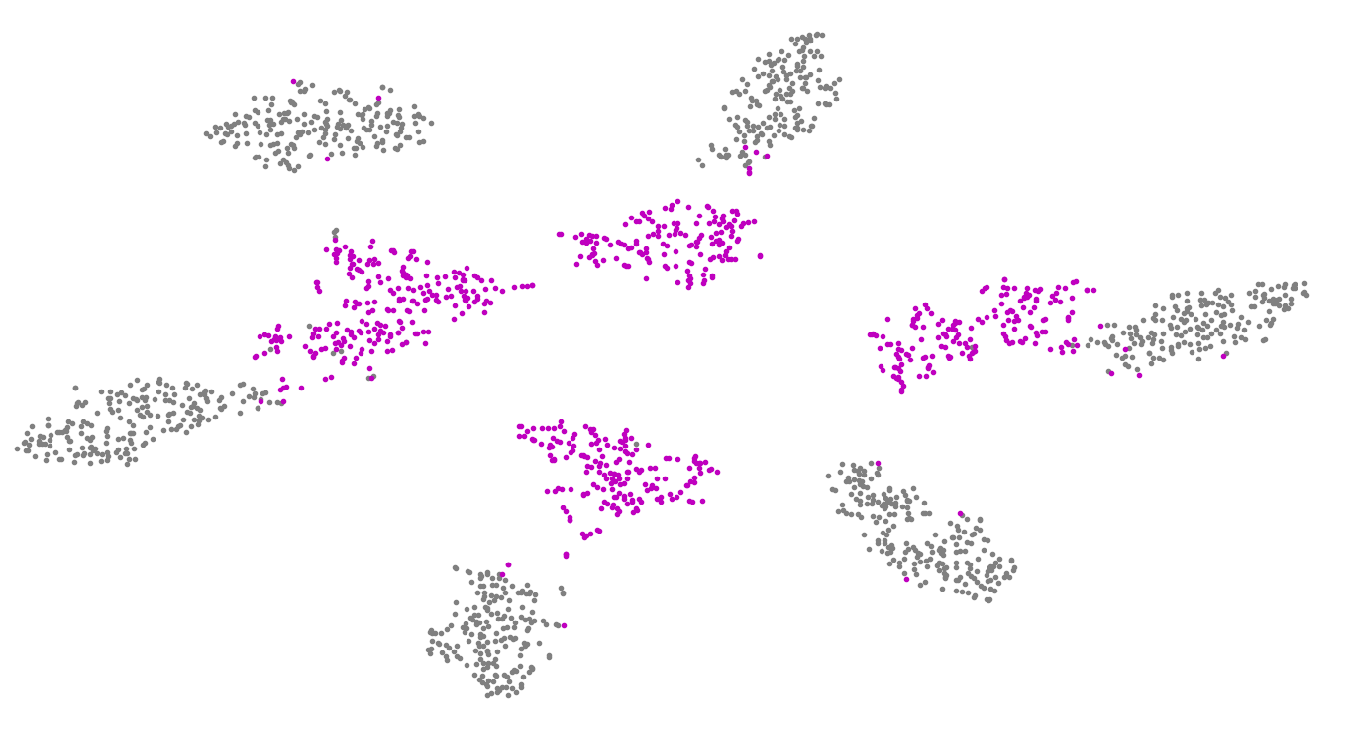}  }
\subfigure[AMPF++] {   \label{AMPF++_inner} 
\includegraphics[width=0.65\columnwidth]{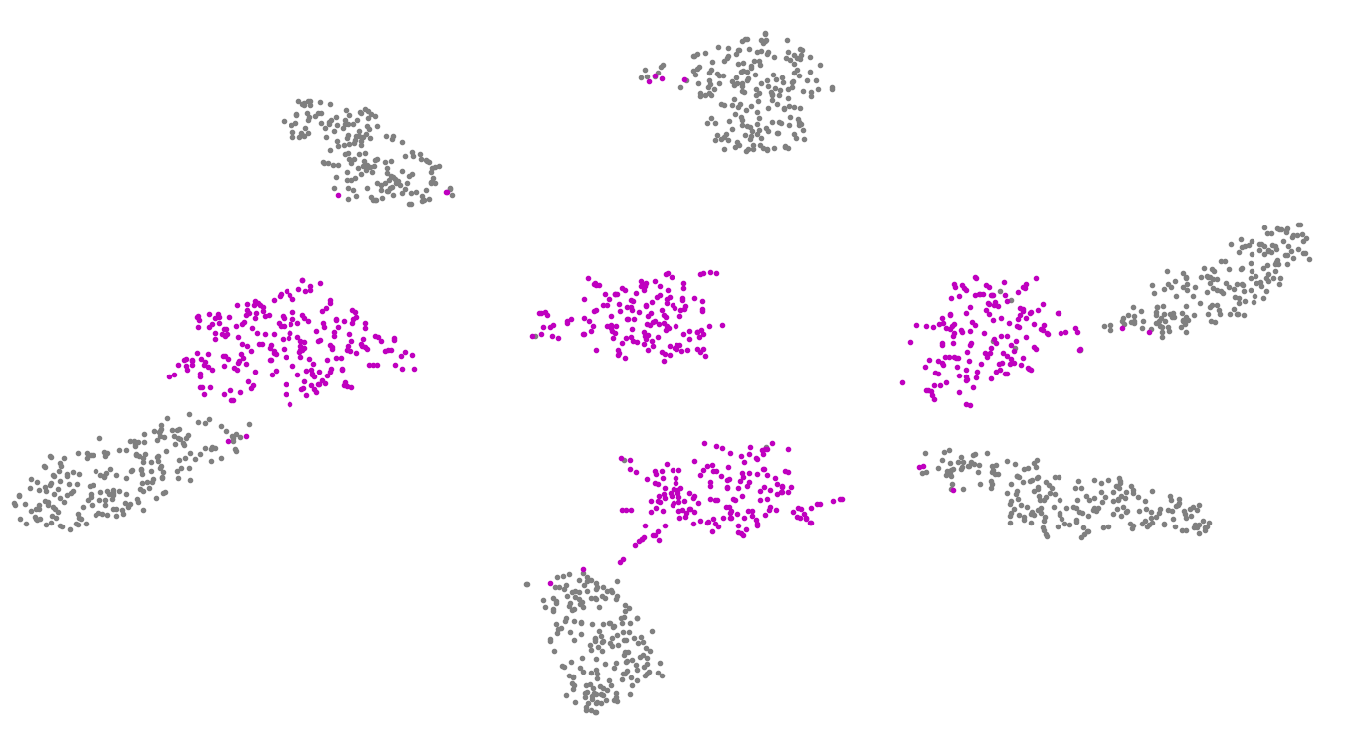}  }

\subfigure[MPF(open set)] {  \label{MPF_outer}  
\includegraphics[width=0.65\columnwidth]{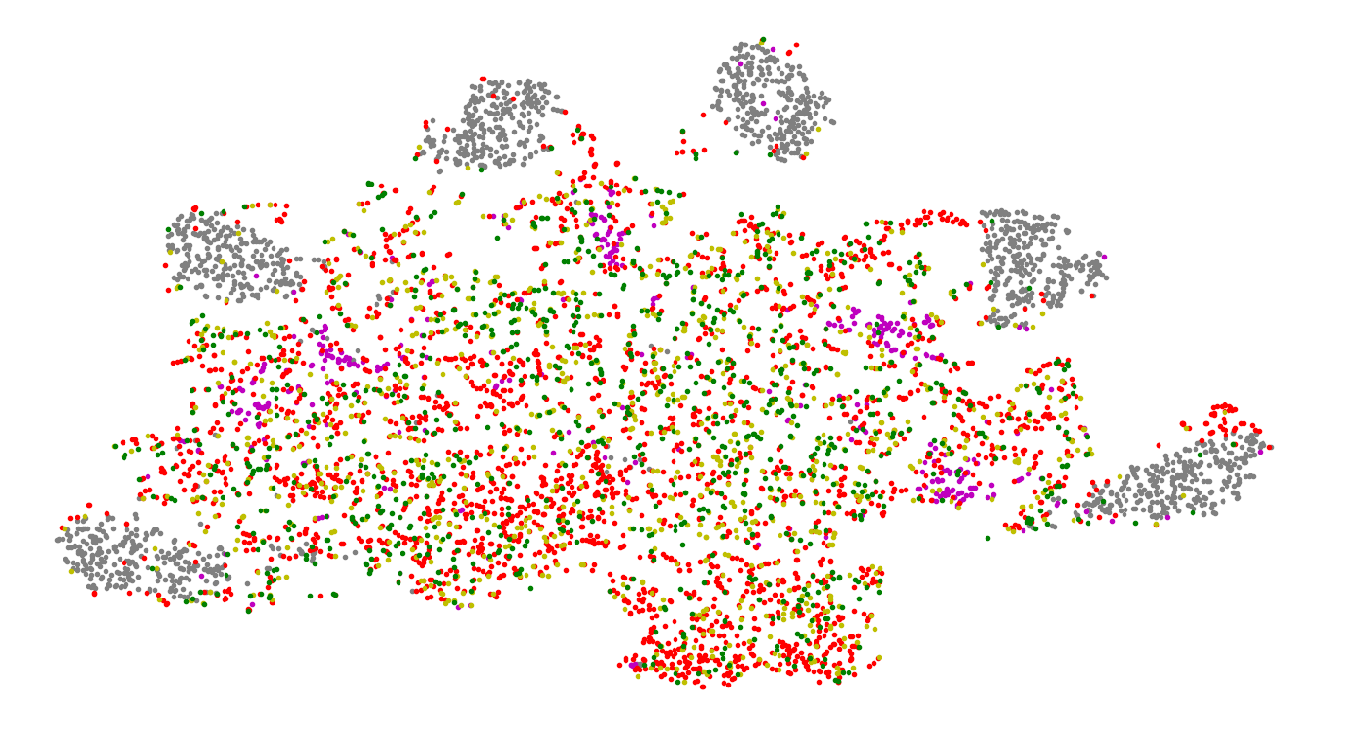}  }
\subfigure[AMPF(open set)] {   \label{AMPF_outer} 
\includegraphics[width=0.65\columnwidth]{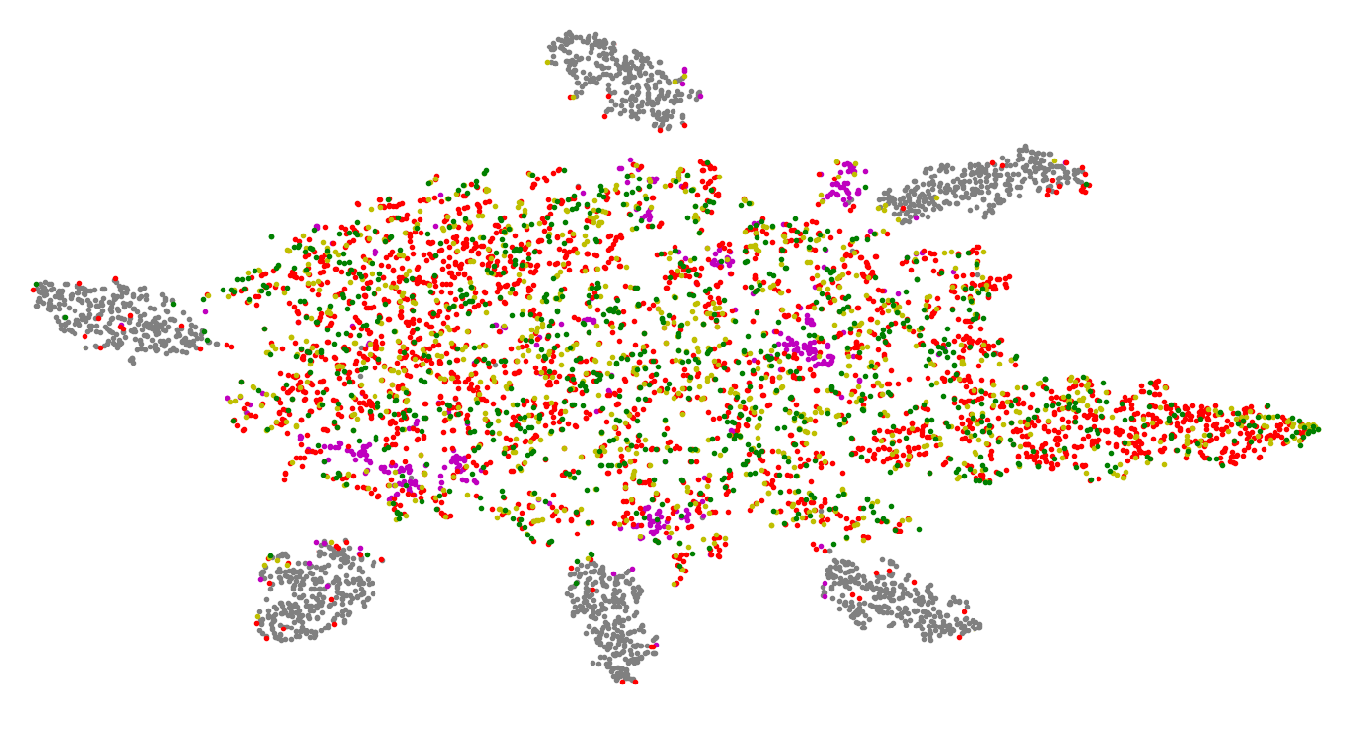}  }
\subfigure[AMPF++(open set)] {   \label{AMPF++_outer} 
\includegraphics[width=0.65\columnwidth]{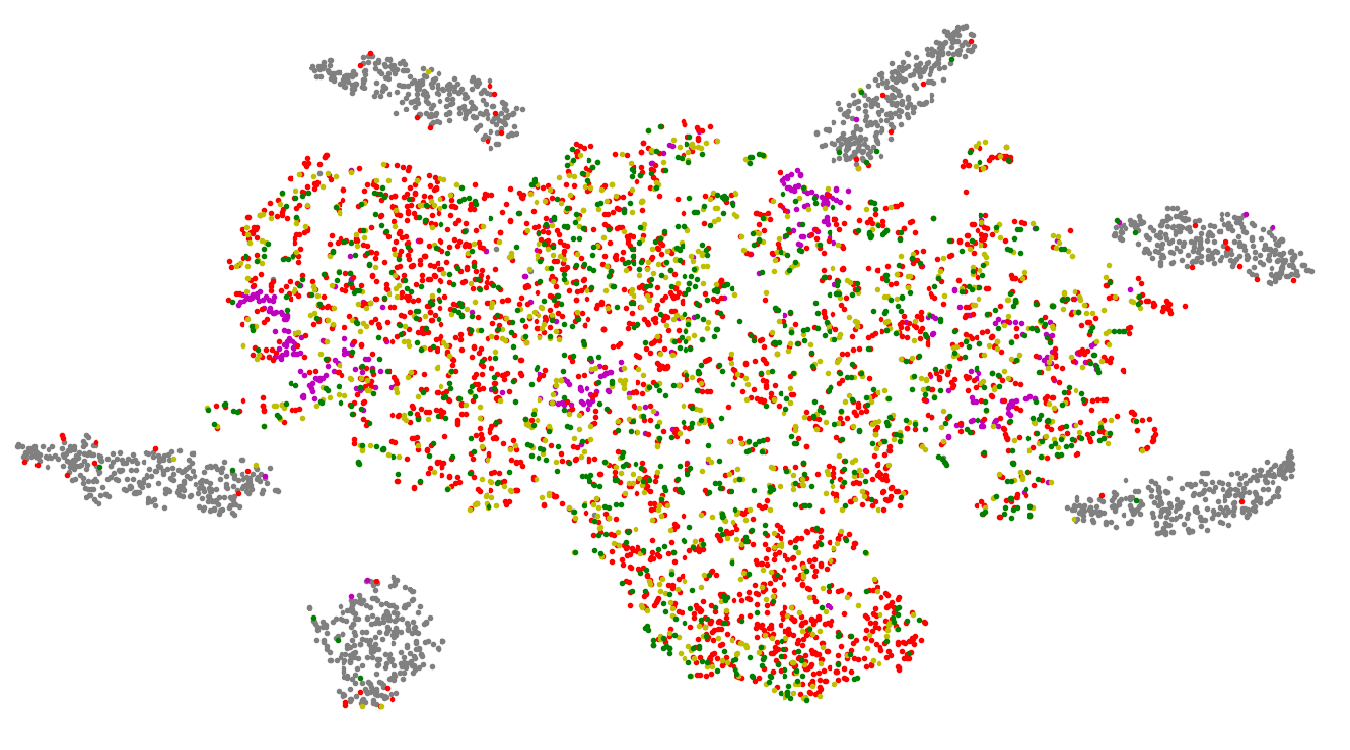}  }

\caption{\textbf{The t-SNE visualization results on known and unknown classes}. MNIST (gray) is used as a known class, and SVHN (red), CIFAR10 (green) and CIFAR100 (yellow) are used as unknown classes for open set evaluation. Different from Figs. \ref{the 1st fig} and \ref{the 7th fig}, known classes only have $6$ categories rather than $10$ here, and the pink dots represent the $4$ remaining unknown classes of MNIST. The first row in the figure shows the visualization results on the known and unknown classes of MNIST under our models, while the unknown classes in the second row are more complex.}
\label{the 8th fig}
\end{figure*}

\subsubsection{Visualization of 2D Features}
According to Fig. \ref{the 7th fig}, all three models can effectively reduce the empirical risk. Without the optimization of loss function $L_o$, as shown in Fig. \ref{MPF_0_vis}, there is a large overlap in the embedding features between the known and unknown classes. In addition, the cluster of known classes is not compact, which proves the effectiveness of the margin constraint term $L_o$. Due to the different network structures used, the visualization results of Fig. \ref{MPF open set visualization} and \ref{MPF_1_vis} are slightly different.

Compared with MPF, AMPF and AMPF++ introduce an adversarial optimization strategy, and they can map adversarial samples to the edge region of the open space. As shown in Fig. \ref{AMPF_vis} and \ref{AMPF++_vis}, some KMNIST and SVHN data are considered similar to the known classes, and they are mapped around the known classes. In addition, the embedding features of unknown classes are effectively suppressed within a small range in the center of the open space, which is also significant to reducing the open space risk. It is possible that CIFAR10 and CIFAR100 are not very similar to the known class MNIST, and thus, few data of these two categories are mapped to the surrounding area.

\subsubsection{t-SNE Visualization}
According to the first row in Fig. \ref{the 8th fig}, all of our models can distinguish known and unknown classes of MNIST effectively. In particular, none of the four unknown classes overlaps with the known classes, and all categories of MNIST form complete clusters. However, the network parameters are not updated during this test phase at all. Therefore, our models have some class-incremental learning ability.

In the second row of Fig. \ref{the 8th fig}, the t-SNE visualization results on the open set are shown. As shown in these figures, the six known classes'' clusters can be seen, and the visualization effect improves with increasing complexity of the model: the degree of overlap between known and unknown classes is decreasing, and the classification boundary between known and unknown classes is becoming clearer. Hence, these results prove that the performance of these three models is steadily improving.

\subsection{Experiments on ImageNet} \label{section on ImageNet}

\begin{table*}
\begin{center}
\caption{The OSR experiment results on ImageNet and the best results are indicated in bold. We reproduce the softmax, GCPL, RPL and ARPL except CPN(we just copy its results from \cite{GCPL-journal}).}
\begin{tabular}{cccccccccc}
\toprule
\multirow{2}*{Method} & \multicolumn{3}{c}{ImageNet-100} & \multicolumn{3}{c}{ImageNet-200}& \multicolumn{3}{c}{ImageNet-1000} \\
\cmidrule(lr){2-4}\cmidrule(lr){5-7}\cmidrule(lr){8-10} & ACC(\%) & AUROC(\%) & OSCR(\%) & ACC(\%) & AUROC(\%) & OSCR(\%) & ACC(\%) & AUROC(\%) & OSCR(\%) \\ 
\midrule
Softmax                   & 80.0 & 89.1 & 75.3 & 83.1 & 92.6 & 80.5 & 69.6 & 48.2 & 42.4\\ 
GCPL\cite{GCPL-conference}& 66.8 & 80.6 & 59.9 & 65.6 & 85.5 & 60.8 & 34.0 & 55.3 & 25.3\\
RPL\cite{RPL}             & 75.0 & 93.9 & 70.8 & 66.2 & 91.7 & 62.3 & 58.8 & 57.5 & 41.9\\
ARPL\cite{ARPL}           & 80.5 & 94.4 & 78.2 & 82.3 & 94.9 & 79.6 & 70.2 & 60.0 & 48.9\\
CPN\cite{GCPL-journal}    & \textbf{86.2} & 82.3 & -    & 82.2 & 79.6 & - & - & - & -\\
\midrule
MPF   &80.8 &\textbf{94.6} &\textbf{78.7}&\textbf{83.8} &\textbf{95.6} &\textbf{81.3} & \textbf{70.4} & \textbf{61.4} &\textbf{49.4}  \\
\bottomrule
\end{tabular} \label{ImageNet table}
\end{center}
\end{table*}

\begin{figure*}[ht]
\centering
\subfigure[K=2(test set)] {\label{K=2_test}     
\includegraphics[width=0.38\columnwidth]{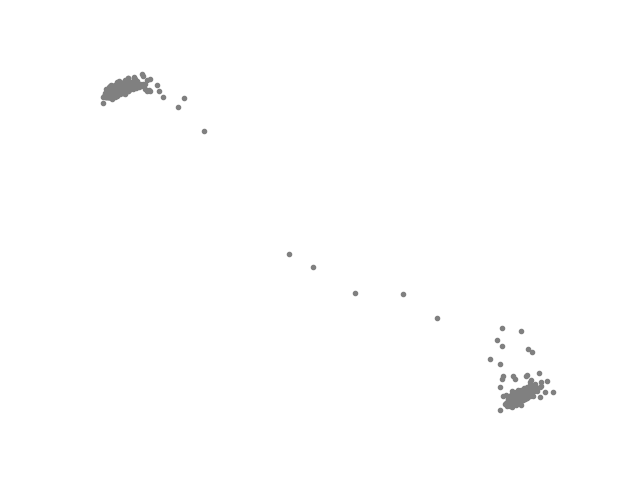}  }
\subfigure[K=3(test set)] {\label{K=3_test}      
\includegraphics[width=0.38\columnwidth]{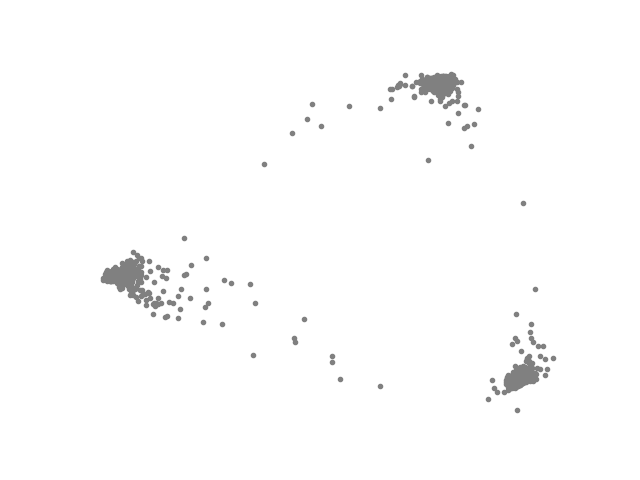}  }
\subfigure[K=5(test set)] {\label{K=5_test}     
\includegraphics[width=0.38\columnwidth]{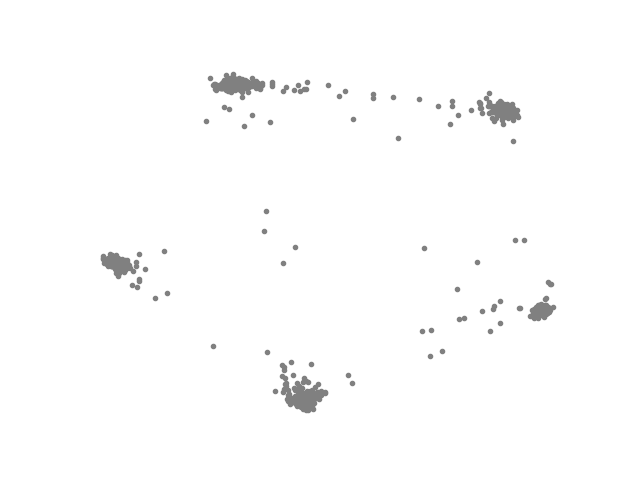}  }
\subfigure[K=8(test set)] { \label{K=8_test}      
\includegraphics[width=0.38\columnwidth]{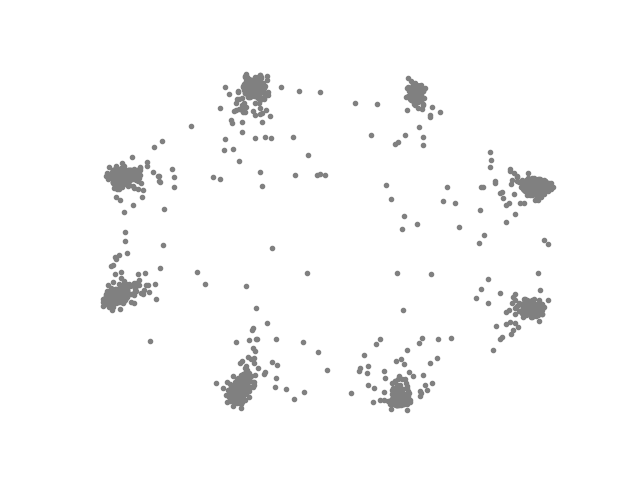}   } 
\subfigure[K=10(test set)] { \label{K=10_test}      
\includegraphics[width=0.38\columnwidth]{MPFLoss_visualize_test_loader.png}   }

\subfigure[K=2(open set)] {\label{K=2_open}    
\includegraphics[width=0.38\columnwidth]{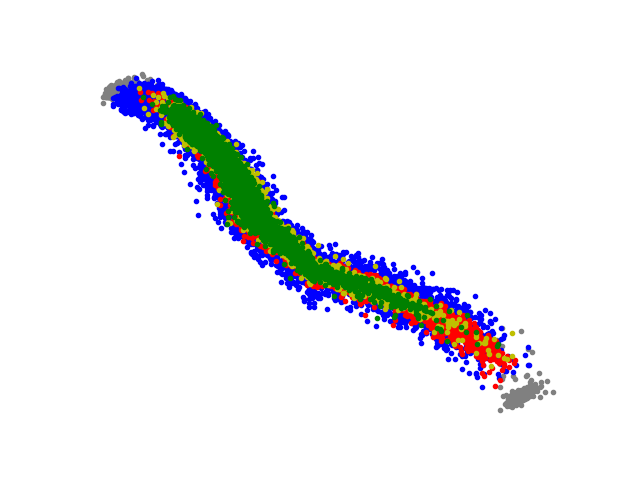}  }
\subfigure[K=3(open set)] {\label{K=3_open}      
\includegraphics[width=0.38\columnwidth]{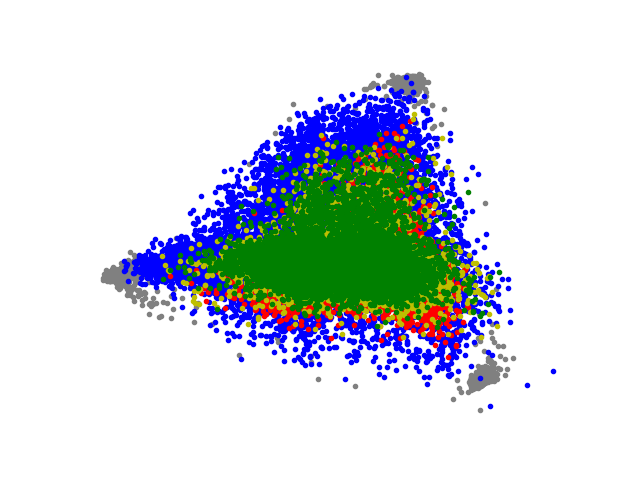}  }
\subfigure[K=5(open set)] {\label{K=5_open}     
\includegraphics[width=0.38\columnwidth]{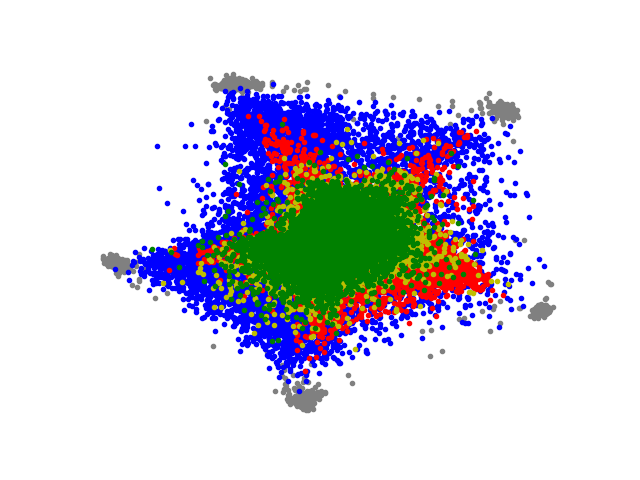}  }
\subfigure[K=8(open set)] {\label{K=8_open}     
\includegraphics[width=0.38\columnwidth]{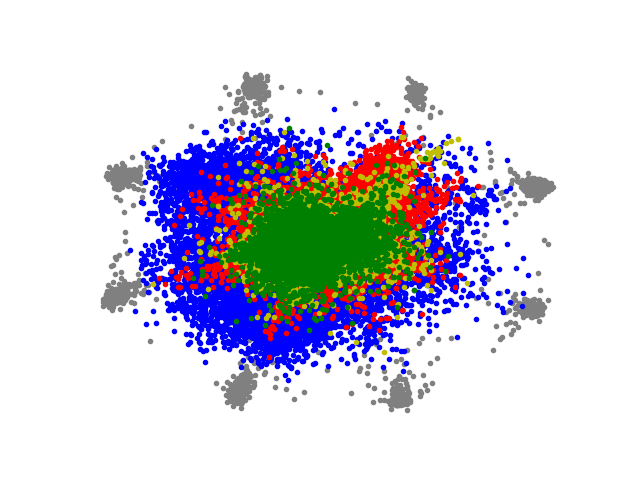}  } 
\subfigure[K=10(open set)] {\label{K=10_open}      
\includegraphics[width=0.38\columnwidth]{MPFLoss_visualize_open_set.png}  }

\caption{\textbf{The visualization results of LENET++ on known and unknown classes with different numbers of known classes}\cite{objectosphere}. Similar to Fig. 1, MNIST (gray) is used as a known class, and KMNIST (blue), SVHN (red), CIFAR10 (green) and CIFAR100 (yellow) are used as unknown classes for open set evaluation. When $K$ is not equal to $10$, the corresponding $K$ known classes are randomly selected from MNIST. In addition, Fig. \ref{K=10_test} and \ref{K=10_open} are exactly the same as Fig. \ref{MPF test data visualization} and \ref{MPF open set visualization}.}
\label{MPF with different K}
\end{figure*}

\begin{figure*}[ht]
\centering
\subfigure[K=2] {\label{K=2_train}     
\includegraphics[width=0.38\columnwidth]{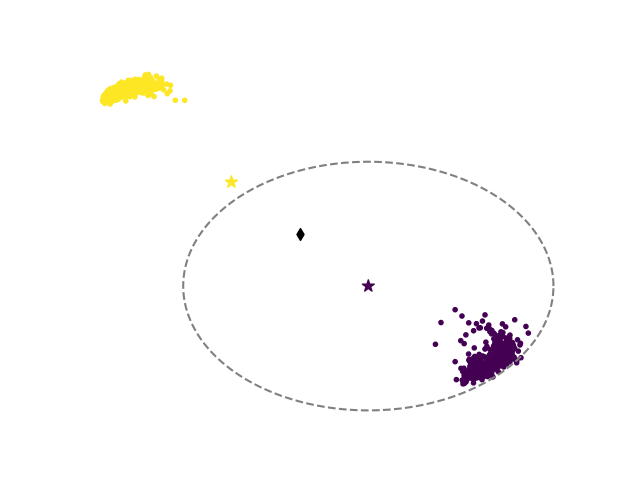}  }
\subfigure[K=3] {\label{K=3_train}      
\includegraphics[width=0.38\columnwidth]{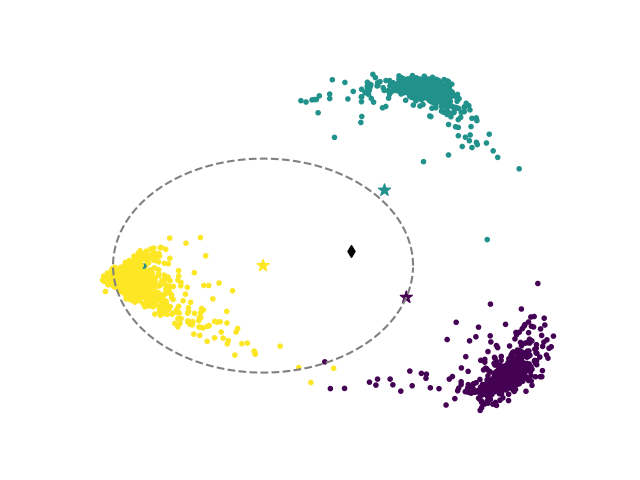}  }
\subfigure[K=5] {\label{K=5_train}     
\includegraphics[width=0.38\columnwidth]{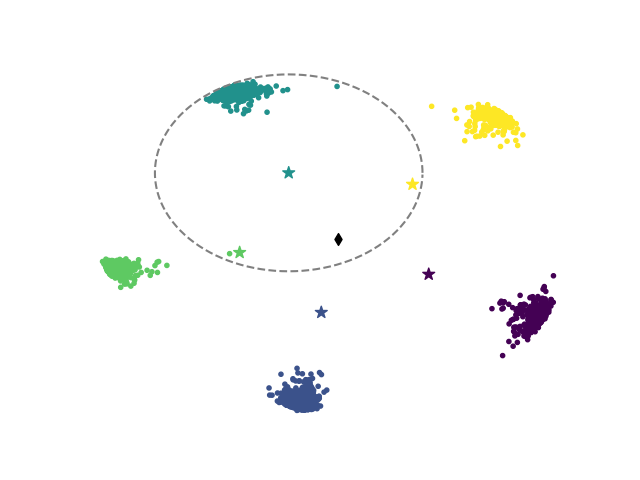}  }
\subfigure[K=8] { \label{K=8_train}      
\includegraphics[width=0.38\columnwidth]{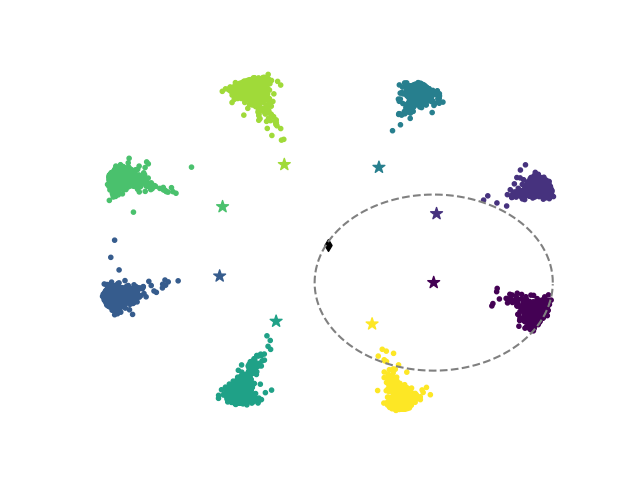}   } 
\subfigure[K=10] {\label{K=10_train}      
\includegraphics[width=0.38\columnwidth]{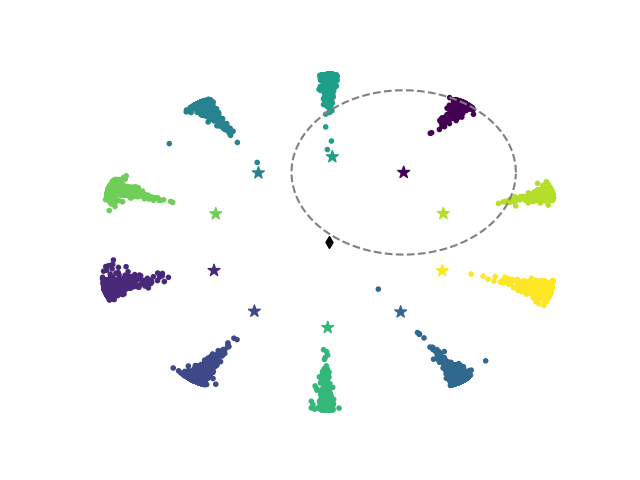}   }

\caption{\textbf{The visualization results of LENET++ on training data with different numbers of known classes}\cite{objectosphere}. In every subfigure, different colored dots represent different classes in MNIST, the star corresponds to the prototype of the each class, the black diamond represents the center $O_c$, and the gray dashed line represent the radius $R$ for a randomly selected prototype.}
\label{MPF on training data with different K}
\end{figure*}

Prior to this paper, there were few performance comparison results between different OSR methods on large data sets. To better prove the effectiveness of the proposed method, experiments are performed here on ImageNet data sets\cite{TinyImageNet}, which is much larger and more difficult. ImageNet includes $1000$ classes with $1,281,167$ training images and $50,000$ validation images.

Firstly, the first $100$ and $200$ classes are selected as known classes, and the remainder are regarded as unknown classes, which are denoted as "ImageNet-100" and "ImageNet-200", respectively. ResNet$50$ is used as the classifier network in this experiment, and it is trained on training images from known classes and tested on all validation images. Moreover, all of $1000$ classes in ImageNet are used as known classes, and ImageNet-O is used as unknown classes\cite{ImageNet-O}. ResNet$18$ instead of ResNet$50$ is used in this experiment. As in the previous experiments, ACC (the closed set accuracy on the known classes data), AUROC, and OSCR are selected to evaluate the performance of the model.

Tabs. \ref{AUROC table} and \ref{OSCR table} show that the performance of MPF is not much better than that of ARPL when the test datasets are not large. However, as shown in Tab. \ref{ImageNet table}, the performance of MPF is higher than that of ARPL in all three experiments. Moreover, MPF performs better than traditional softmax and some other methods(except CPN in ImageNet-100), which shows the excellent scalability of the proposed method on larger-scale datasets.

\subsection{MPF With Different Numbers of Known Classes}
For a more complete presentation of the MPF model, this section shows the visualization results of the model with different numbers of known classes. In this paper, the number of known classes is denoted as $K$, and it is set to $2$, $3$, $5$, $8$ and $10$.

As shown in the first row of Fig. \ref{MPF with different K}, MPF tends to distribute known classes symmetrically in the feature space. This distribution pattern not only is consistent with human aesthetic preferences but also is the optimal way to utilize the feature space. In addition, MPF can effectively reduce the empirical risk regardless of the value of $K$.

When a large number of unknown classes is added into the test set, as shown in the second row of Fig. \ref{MPF with different K}, MPF can effectively reduce the two risks. It can also be seen that KMNIST (blue) has the greatest overlap with the known classes, and this overlap decreases with the increase in $K$. Therefore, these results also prove that MPF is suitable for OSR with a large number of known classes, as shown in section \ref{section on ImageNet}.

\subsection{The Distance Setting in Our Models}

This subsection uses visualization to verify the validity of the distance setting in this article. As shown in Fig. \ref{MPF on training data with different K}, no matter what the number of known classes is, MPF can always complete clustering effectively. As the loss function of MPF is designed, the known classes features are covered by the radius $R$. Similar to Fig. \ref{distance setting}, embedding features, prototypes and the center of feature space(because of symmetry, the origin of coordinates is always approximately equal to the center $O_c$) are always approximately in a straight line. Moreover, as the number of known classes increases, the degree of this three-point one-line approximation becomes higher. This distance setting avoids feature clustering as shown in Fig. \ref{GCPL test data visualization}, so it will help the model reduce the open space risk.

\section{Conclusion}
In a sense, reducing the empirical risk means moving from "known" to "known", while reducing the open space risk means moving from "known" to "unknown". Due to the complexity and uncertainty of the "unknown", it is obviously much more difficult to reduce the open space risk. For this reason, most OSR studies are devoted to effectively reducing the open space risk, as is the case in this paper. The MPF, AMPF and AMPF++ proposed in this paper are increasingly effective in reducing the open space risk.

There are two main reasons for the excellent OSR performance of AMPF and AMPF++:
\begin{enumerate}
\item These models generate a large amount of unknown data, and the classifier has improved its ability to identify "unknown" because it has seen generated unknown data in the training phase;
\item With the "adversarial motion" of margin constraint radius $R$, the differential mapping ability of the classifier to the known and unknown classes is greatly enhanced; when $R$ increases, the clustering of known classes is strengthened and corresponds to a reduction in the empirical risk; and when $R$ is reduced, the classifier maps the unknown classes to the edge region of the open space, which corresponds to a reduction in the open space risk.
\end{enumerate}

This novel and adversarial optimization strategy designed in this paper is reflected not only in the generation of adversarial samples but also in the "adversarial motion" of the margin constraint radius $R$. With the motion of $R$, both risks of the model are reduced alternately, and this alternate reduction mode is more effective than the single reduction mode, such as \cite{GCPL-conference,RPL,ARPL} and \cite{GCPL-journal}.

\ifCLASSOPTIONcaptionsoff
  \newpage
\fi

\bibliographystyle{unsrt}
\bibliography{ref}

\begin{IEEEbiography}[{\includegraphics[width=1in,height=1.25in,clip,keepaspectratio]{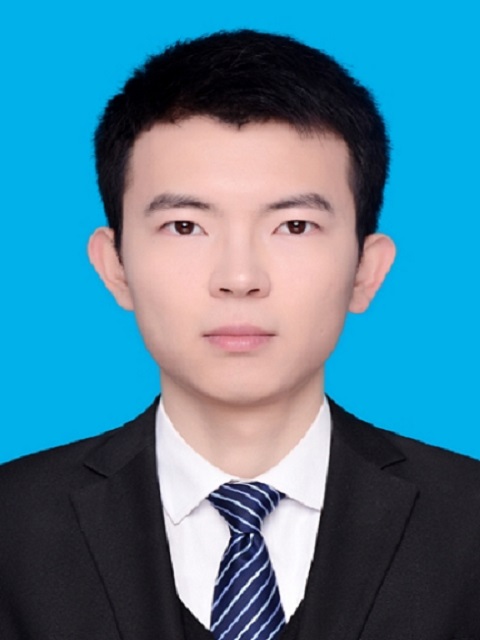}}]{Ziheng Xia}
Ziheng Xia received B.S. degree from Peking University in Nuclear Physics in $2015$, and received M.Eng. degree from Northwest Institute of Nuclear Technology in Nuclear Technology and Application in $2017$. And now, He is studying for his Ph.D. degree in signal processing at Xidian University. His research interests include radar automatic target recognition, pattern recognition and machine learning.
\end{IEEEbiography}

\begin{IEEEbiography}[{\includegraphics[width=1in,height=1.25in,clip,keepaspectratio]{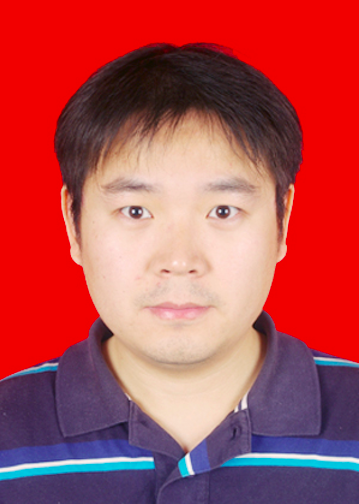}}]{Penghui Wang}
Penghui Wang received the B.S. degree in communication engineering from National University of Defense Technology (NUDT), Changsha, China, in 2005 and Ph.D. degree in signal processing from Xidian University, Xi’an, China, in 2012. He is now an associate professor at the National Laboratory of Radar Signal Processing, Xidian University. His research interests include radar signal processing and automatic target recognition.
\end{IEEEbiography}

\begin{IEEEbiography}[{\includegraphics[width=1in,height=1.25in,clip,keepaspectratio]{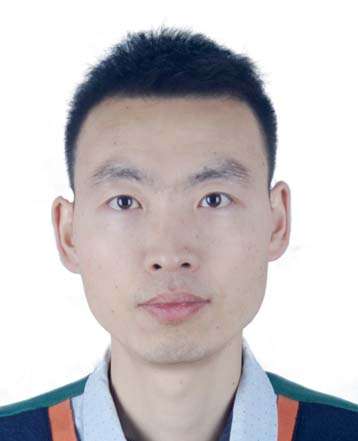}}]{Ganggang Dong}
Ganggang Dong received M.S. degree and Ph.D. degree in information and communication engineering from the National University of Defense Technology, Changsha, China, in $2012$ and $2016$. Dr. Dong was the winner of $2017$ Excellent Doctoral Dissertations of CIE (Chinese Institute of Electronics). He authored more than $30$ scientific papers in peer-reviewed journals and conferences, including IEEE Cybernetics, IEEE TIP, IEEE TGRS, IEEE JSTARS, IEEE GRSL, and IEEE SPL. His research interests include target detection and recognition.
\end{IEEEbiography}

\begin{IEEEbiography}[{\includegraphics[width=1in,height=1.25in,clip,keepaspectratio]{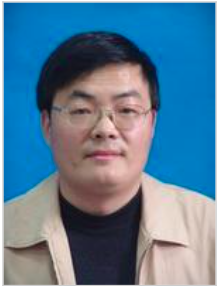}}]{Hongwei Liu}
Hongwei Liu received the B.Eng. degree from Dalian University of Technology in electronic engineering in $1992$, and the M.Eng. and Ph.D degrees in electronic engineering from Xidian University, Xi’an, China, in $1995$ and $1999$, respectively. He is currently the Director and a Professor with the National Laboratory of Radar Signal Processing, Xidian University. His research interests include radar automatic target recognition(RATR), radar signal processing, and adaptive signal processing.
\end{IEEEbiography}

\end{document}